%% file: main.tex
\pdfoutput=1
\documentclass{article}

\usepackage[final]{corl_2026}

\usepackage{amsthm}
\usepackage{amsmath}
\usepackage{amsfonts}
\usepackage{amssymb}
\usepackage{mathtools}
\usepackage{graphicx}
\usepackage{wrapfig}
\usepackage{enumitem}
\usepackage{booktabs}
\usepackage{xcolor}

\usepackage[font=footnotesize]{caption}
\usepackage{algorithm}
\usepackage{algpseudocode}

\title{Flow Control: Steering Vision--Language--Action Models with Simple Real-Time Inputs}

\author{
  Jonathan C. Kao \\
  Departments of Electrical and Computer Engineering, \\
  Computer Science, and Neurobiology,\\
  University of California, Los Angeles \\
  \texttt{kao@seas.ucla.edu} \\
  \And
  Jason Chan \\
  Department of Computer Science \\
  University of California, Los Angeles \\
  \texttt{jasontchan@ucla.edu} \\
  \AND
  Andy Wang \\
  Department of Computer Science \\
  University of California, Los Angeles \\
  \texttt{andywang0321@ucla.edu} \\
}

\hypersetup{
  pdftitle={Flow Control: Steering Vision-Language-Action Models with Simple Real-Time Inputs},
  pdfauthor={Jonathan C. Kao, Jason Chan, Andy Wang}
}

\begin{document}

\maketitle

\vspace{-0.6em}
\input{0_abstract}


\input{math_preamble}

\input{1_introduction}
\input{2_related}
\input{3_methods}
\input{4_experiments}
\input{5_discussion}

\clearpage
\acknowledgments{
This work was supported by NIH DP1HD121548, DP2NS122037, and the UCLA-Amazon Science Hub award (all to J.C.K.).

\textbf{Author contributions.}
J.C.K. conceived and implemented flow control, conducted the experiments and user study, performed the analyses, generated the figures, and wrote the paper.
J.C. wrote code to operate the robotic arm.
J.C. and A.W. participated in paper review.

\textbf{Competing interests.}
J.C.K. is the inventor of intellectual property owned by Stanford University that has been licensed to Blackrock Neurotech and Neuralink Corp.
J.C.K. has a provisional patent application related to AI copilots for brain-computer interfaces that is owned by the Regents of the University of California.
J.C.K. is a co-founder of Luke, is on its Board of Directors and has a financial interest in it.
The other authors declare no competing interests.
}

\bibliography{references}

\appendix
\input{appendix}

\end{document}

%% file: 0_abstract.tex
\begin{abstract}

We introduce flow control of vision-language-action (VLA) models, a simple and effective way to steer VLA actions in real time through generic inputs, such as a keyboard.
This method can be used out-of-the-box and does not require retraining or fine-tuning VLAs.
It enables relatively crude user inputs to steer a VLA to align with user intent.
Flow control writes the user input into the initial condition of the VLA's flow matching action expert.
Because the flow is deterministic, integration retains information about the user input while moving it toward the action distribution learned during training, so that the generated actions reflect the user's intent (high fidelity) and stay close to the VLA's own unsteered actions (high conformity).
We demonstrate that flow control has many desirable properties: (1) flow control accurately and responsively steers robot actions with user inputs, (2) it is robust to suboptimal user inputs, (3) it enables users to steer VLAs to achieve significantly higher success rates and faster task completion, and (4) fine-tuning a VLA on flow control trajectories improves the autonomous policy.
Together, these results provide a simple and intuitive way for users to help steer VLA actions, increasing task performance.
Project page and videos: \url{https://jcykao.github.io/flowcontrol/}.

\end{abstract}

%% file: math_preamble.tex
\newcommand{\0}{\mathbf{0}}
\newcommand{\1}{{(1)}}
\newcommand{\2}{{(2)}}
\newcommand{\3}{{(3)}}
\renewcommand{\a}{\mathbf{a}}
\newcommand{\A}{\mathbf{A}}
\renewcommand{\b}{\mathbf{b}}
\newcommand{\bias}{\textrm{bias}}
\newcommand{\B}{\mathbf{B}}
\newcommand{\BN}{\textrm{batch-norm}}
\renewcommand{\c}{\mathbf{c}}
\renewcommand{\c}{\mathbf{c}}
\newcommand{\C}{\mathbf{C}}
\newcommand{\CE}{\textrm{CE}}
\newcommand{\cov}{\mathbf{cov}}
\renewcommand{\d}{\partial}
\newcommand{\D}{\mathcal{D}}
\newcommand{\data}{{\textrm{data}}}
\newcommand{\Dsup}{{(D)}}
\newcommand{\E}{\mathbb{E}}
\newcommand{\e}{\epsilon}
\newcommand{\eps}{\varepsilon}
\newcommand{\f}{\mathbf{f}}
\newcommand{\G}{\mathcal{G}}
\newcommand{\Gsup}{{(G)}}
\newcommand{\g}{\mathbf{g}}
\newcommand{\grad}{\nabla}
\newcommand{\gradlogpi}{\grad_\theta \log \pi_\theta}
\newcommand{\gradlogp}{\grad_\theta \log p_\theta}
\renewcommand{\H}{\mathbf{H}}
\newcommand{\h}{\mathbf{h}}
\newcommand{\hinge}{\textrm{hinge}}
\newcommand{\hmu}{\hat{\mu}}
\newcommand{\htheta}{\hat{\theta}}
\newcommand{\hx}{\hat{\x}}
\newcommand{\hy}{\hat{y}}
\newcommand{\I}{\mathbf{I}}
\newcommand{\ib}{\mathbf{i}}
\newcommand{\intext}{{\textrm{in}}}
\newcommand{\Ind}{\mathbb{I}}
\newcommand\ind{\protect\mathpalette{\protect\independenT}{\perp}}
\def\independenT#1#2{\mathrel{\rlap{$#1#2$}\mkern2mu{#1#2}}}
\newcommand{\J}{\mathbf{J}}
\newcommand{\JS}{\textrm{JS}}
\renewcommand{\j}{{(j)}}
\newcommand{\KL}{\textrm{KL}}
\renewcommand{\L}{\mathcal{L}}
\newcommand{\Lvae}{\L_{\textrm{vae}}}
\newcommand{\leftV}{\left\Vert}
\newcommand{\m}{{(m)}}
\newcommand{\maxout}{\textrm{maxout}}
\newcommand{\model}{{\textrm{model}}}
\newcommand{\MSE}{\textrm{MSE}}
\newcommand{\N}{\mathcal{N}}
\newcommand{\nab}{\nabla}
\newcommand{\new}{{\textrm{new}}}
\renewcommand{\o}{\mathbf{o}}
\renewcommand{\O}{\mathbf{O}}
\newcommand{\out}{{\textrm{out}}}
\newcommand{\opt}{{\textrm{opt}}}
\newcommand{\p}{\mathbf{p}}
\newcommand{\pad}{\texttt{pad}}
\newcommand{\pmodel}{p_{\mathrm{model}}}
\newcommand{\pdata}{p_{\mathrm{expert}}}
\newcommand{\pinit}{p_{\mathrm{init}}}
\newcommand{\pool}{\texttt{pool}}
\newcommand{\q}{\mathbf{q}}
\newcommand{\rec}{{\textrm{rec}}}
\newcommand{\relu}{\textrm{ReLU}}
\newcommand{\rightV}{\right\Vert}
\newcommand{\range}{\textrm{range}}
\renewcommand{\r}{\mathbf{r}}
\newcommand{\R}{\mathbb{R}}
\newcommand{\calR}{\mathcal{R}}
\newcommand{\s}{\mathbf{s}}
\renewcommand{\S}{\mathbf{S}}
\newcommand{\Si}{\Sigma^{-1}}
\newcommand{\sign}{\textrm{sign}}
\newcommand{\softmax}{\textrm{softmax}}
\newcommand{\softplus}{\textrm{softplus}}
\newcommand{\stride}{\texttt{stride}}
\newcommand{\tausimp}{\tau \sim p_\theta(\tau)}
\newcommand{\SE}{\textrm{SE}}
\newcommand{\tA}{\mathsf{A}}
\newcommand{\thetap}{\theta^\prime}
\newcommand{\tJ}{\tilde{J}}
\newcommand{\tp}{t^\prime}
\newcommand{\tr}{\textrm{tr}}
\newcommand{\tX}{\mathsf{X}}
\newcommand{\tx}{\tilde{\x}}
\renewcommand{\u}{\mathbf{u}}
\newcommand{\user}{\textrm{user}}
\newcommand{\U}{\mathbf{U}}
\renewcommand{\v}{\mathbf{v}}
\newcommand{\w}{\mathbf{w}}
\newcommand{\W}{\mathbf{W}}
\newcommand{\var}{\textrm{var}}
\newcommand{\V}{\mathbf{V}}
\newcommand{\X}{\mathbf{X}}
\newcommand{\x}{\mathbf{x}}
\newcommand{\xdes}{\mathbf{x}_\textrm{des}}
\newcommand{\xor}{\textrm{xor}}
\newcommand{\Xdes}{\mathcal{X}_{\textrm{desired}}}
\newcommand{\Xrej}{\mathcal{X}_{\textrm{reject}}}
\newcommand{\y}{\mathbf{y}}
\newcommand{\Y}{\mathbf{Y}}
\newcommand{\z}{\mathbf{z}}
\newcommand{\Z}{\mathbf{Z}}
\newcommand{\calA}{\mathcal{A}}
\newcommand{\calD}{\mathcal{D}}
\newcommand{\calE}{\mathcal{E}}
\newcommand{\calM}{\mathcal{M}}
\newcommand{\calO}{\mathcal{O}}
\newcommand{\calS}{\mathcal{S}}
\newcommand{\calT}{\mathcal{T}}

%% file: 1_introduction.tex
\section{Introduction}

Vision-language-action (VLA) models map camera observations and natural language commands to robot actions, achieving state-of-the-art performance on many tasks \cite{Black2024-zr, Intelligence2025-ht, Brohan2023-er, Driess2023-py, Szot2024-rv, Gemini-Robotics-Team2025-pm, Kim2024-xw, Intelligence2026-xj}.
But VLAs also remain unreliable in many ways:
they may follow language poorly \cite{Chen2026-go, Gao2025-yn, Intelligence2026-xj, Lee2025-uo}, fail to generalize to novel objects and scenes \cite{Intelligence2025-ht, Kim2024-xw}, and, like many imitation-learning based policies, drift into out-of-distribution (OOD) states where recovery is challenging \cite{Ross2010-vu}.
These shortcomings motivate providing users with a way to steer VLAs as they act.
Existing approaches use language through mid-episode corrections or hierarchical commands \cite{Shi2025-oh, Belkhale2024-ns, Intelligence2026-xj}, as well as trajectory sketches and goal images \cite{Lee2025-uo, Zhao2025-jk, Intelligence2026-xj} to help steer VLA actions.
But language corrections are coarse and intermittent, while sketches and goal images require high bandwidth.
These steering methods furthermore involve additional data collection and fine-tuning.

In this work, we aim to address these limitations by enabling users to steer VLA actions in real time, providing low-bandwidth but precise adjustments and/or corrections.
We also aim to achieve this steering without additional policy training, so that VLAs can be steered out-of-the-box.
Finally, we also aim for this steering interface to be \textit{intuitive and simple}, not requiring significant user learning as in teleoperation.
This goal is summarized in Figure~\ref{fig:fig1}, where simple and generic user inputs (in this study, keyboard inputs) can be injected into a frozen VLA (in this study, $\pi_{0.5}$) to steer VLA robot actions without any retraining.
Although our results focus on user steering, a vision-language model (VLM) can provide the same inputs to steer VLA actions (Appendix~\ref{app:vlm}).

\input{fig1}

A critical observation we leverage is that several state-of-the-art VLAs use generative action experts, such as a flow matching head in $\pi_{0.5}$ \cite{Intelligence2025-ht}.
These generative policies transform noise -- conditioned on camera inputs, language, state, and attention tokens derived from these inputs -- into continuous actions \cite{chi2023diffusion, Chen2024-rq, Black2024-zr, Intelligence2025-ht}.
We call the distribution of these continuous actions the ``action expert distribution.''
In this work, we enable users to provide relatively crude steering inputs through a keyboard, such as \{up, down, left, right, forward, backward\} (see Figure~\ref{fig:fig1}).
On their own, these inputs would be too coarse to provide precise and intuitive steering of high degree-of-freedom (DOF) robots.
However, by using the VLA's generative action expert, we show it is possible for the VLA to transform these crude user inputs into high-quality actions that stay close to the action expert distribution.

Our main contribution is \textit{flow control}, a method that injects user intent to steer VLAs so that generated actions are high quality (close to the action expert distribution) and high fidelity (reflecting the user's intent).
Flow control does not require any data labeling, retraining, or fine-tuning and can be used out-of-the-box with VLAs using flow matching action experts.
In a user study with $16$ participants, we demonstrate that when users steer zero-shot $\pi_{0.5}$-DROID with simple keyboard inputs, task success rate and speed of completion significantly increase.
Finally, we show that fine-tuning $\pi_{0.5}$ on flow control trajectories results in significantly higher autonomous policy performance.
We anticipate this method gives users responsive and precise steering over VLAs out-of-the-box.
Beyond improving task outcomes for robot policies, we also anticipate that this approach may have significant impact on assistive robotics, such as brain-computer interfaces, where low-dimensional user inputs can be translated into high-DOF robot actions.

%% file: fig1.tex
\begin{wrapfigure}{r}{0.5\textwidth}
    \vspace{-0.3cm}
    \centering
    \includegraphics[width=\linewidth]{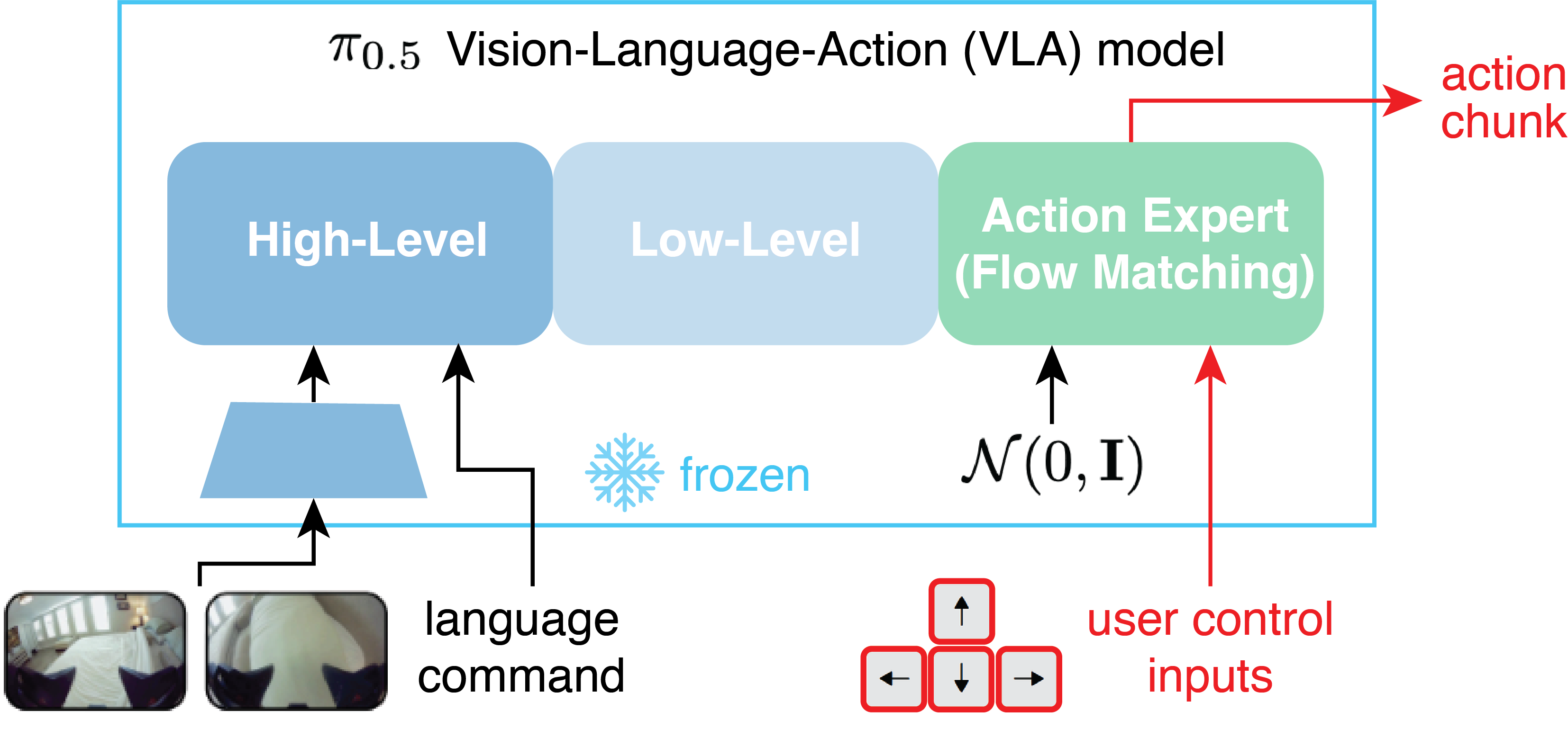}
    \caption{\textbf{Overview}.
    We freeze a VLA with a flow matching action expert.
    Our method (red) steers the VLA by injecting user inputs into its action expert.
    }
    \vspace{-0.3cm}
    \label{fig:fig1}
\end{wrapfigure}

%% file: 2_related.tex
\section{Related Work}

\textbf{Steering robot policies through language, vision, and traces.}
Prior work lets users or high-level modules steer robot policies, differing in the modality of the steering signal and whether it requires retraining.
One common channel is language: hierarchical VLAs decompose tasks into language subtasks a user can issue or correct mid-episode \cite{Belkhale2024-ns, Shi2025-oh}, and a body of work incorporates free-form verbal corrections during execution \cite{Shi2024-pr, cui2023no, Lynch2024-kg, Chisari2023}, with some studies explicitly quantifying and improving a policy's responsiveness to language \cite{Chen2026-go}.
Another method of steering uses spatial or visual goals: hand-drawn 2D trajectories \cite{Sundaresan2024-ua, Gu2023-lo}, editable trajectory traces \cite{Lee2025-uo}, and generated or user-provided subgoal images \cite{Black2023-yw, Zhao2025-jk, Intelligence2026-xj}.
While these interfaces are expressive, they are either coarse, intermittent semantic commands (language) or high-bandwidth spatial inputs (sketches, goal images).
Further, these methods often require additional training on purpose-collected data.
In contrast, flow control uses neither language nor goals: users provide intuitive and simple keyboard inputs in real time, with no additional data collection or training.

\textbf{Inference-time steering of generative policies.}
More related to our work are methods that steer a frozen diffusion or flow policy at inference.
Several guide the sampling process.
Inference-Time Policy Steering (ITPS) \cite{wang2025inferencetimepolicysteeringhuman} biases sampling to align generated actions with objectives such as reaching a point, matching a sketch, or physically correcting an action, computing the gradient of these objectives with respect to the noise and partially stepping along it.
DynaGuide \cite{du2025dynaguidesteeringdiffusionpolices} similarly guides diffusion sampling with the gradient of a metric from a learned dynamics model.
DemoDiffusion \cite{park2025demodiffusiononeshothumanimitation} retargets a single human demonstration into a rough trajectory that a pre-trained policy then refines.
Each steers the policy through its denoising process via objective gradients \cite{wang2025inferencetimepolicysteeringhuman, du2025dynaguidesteeringdiffusionpolices} or a retargeted demonstration prior \cite{park2025demodiffusiononeshothumanimitation}.
Because we target VLAs with flow matching action experts, which are sampled by integrating a deterministic ordinary differential equation (ODE), flow control is significantly simpler: modifying the initial condition of the flow ODE is sufficient to align the VLA with user intent, with no guidance gradients (Section~\ref{subsec:flowcontrol}).
For VLAs that autoregressively generate discrete action tokens, Token Steering \cite{Chan2026-ts} injects low-dimensional user inputs directly into the action tokens at inference.

\textbf{Steering via the initial condition.}
Diffusion steering via reinforcement learning (DSRL) \cite{Wagenmaker2025-hj} also acts on the initial noise rather than guiding the denoising process.
DSRL trains an RL policy $\pi^{\mathcal{W}}(\mathbf{s})$ that replaces the initial-noise distribution $\x_0 \sim \N(\0,\I)$ with $\mathbf{x}_0 \sim \pi^{\mathcal{W}}(\mathbf{s})$, steering the policy from a learned, favorable initial condition.
Our method is simpler: we do not learn a distribution over initial conditions, but set the initial condition directly to a desired action, which the flow matching head then transforms into an action close to the action expert distribution.
Concurrent with our work, Flow Reversal Steering (FRS) \cite{Tang2026-frs} steers flow matching policies through their noise, by integrating the flow in reverse from a reference action and then forward again.
Flow control instead writes the command directly into the initial condition.
In simulations, we found FRS actions were further from the VLA's own actions than flow control actions (Appendix~\ref{app:frs}).

\textbf{Shared autonomy with diffusion policy.}
Prior studies implemented shared autonomy with a diffusion model \cite{yoneda2023noise, Wang2026-ko} to transform user actions into robot actions.
In diffusion-based policies, unit Gaussian noise $\x_0~\sim~\N(\0, \I)$ is reverse diffused to a sample from the action expert distribution, $\x_1 \sim \pdata$.
These methods forward diffuse a user action $\u$ by the proportion $0 \leq \gamma \leq 1$, treat the result as $\x_{1-\gamma}$, and reverse diffuse it, with optional inpainting \cite{Wang2026-ko}, to generate an action sample $\x_1$.
As $\gamma \to 1$, the user action is diffused to unit Gaussian noise, so the sampled action $\x_1$ has \textit{low fidelity} to the user's intent but \textit{high conformity} to $\pdata$.
As $\gamma \to 0$, the action has \textit{high fidelity}, but with fewer reverse diffusion steps $\x_1$ is further from $\pdata$, giving \textit{low conformity}. %
Correctly tuning $\gamma$ is critical.
Flow control avoids this hyperparameter entirely, and because we inject the user's action into the initial condition, it incurs no additional latency to forward- then reverse-diffuse it.
Finally, flow control inputs can be relatively crude because the user action $\u$ is integrated through every step of the flow matching ODE, which moves it toward $\pdata$ while preserving the information it carries (Section~\ref{subsec:ic-information}), yielding both high fidelity and high conformity.

%% file: 3_methods.tex
\section{Methods}

\subsection{Vision-language-action models}

VLAs fine-tune pre-trained vision-language models (VLMs) for robot control, endowing robots with priors learned from large-scale vision and text pre-training \cite{Black2024-zr, Intelligence2025-ht, Brohan2023-er, Driess2023-py, Szot2024-rv, Gemini-Robotics-Team2025-pm, Kim2024-xw}.
In this study, we use the $\pi_{0.5}$ VLA \cite{Intelligence2025-ht}, which fine-tunes the PaliGemma VLM \cite{Beyer2024-uh} on robot state and action chunks.
$\pi_{0.5}$ generates actions with a flow matching action expert, implemented by a downsized $300$M-parameter Gemma transformer \cite{Gemma-Team2024-yf}.
This action expert attends to VLM tokens for camera observations, the language prompt, and the robot state, as well as its own action expert embeddings.
Conditioned on these tokens, the action expert integrates a flow matching ODE, so the generated actions depend on visual, language, and proprioceptive inputs.
We use the VLA $\pi_{0.5}$-DROID, which was fine-tuned on the DROID dataset \cite{Khazatsky2024-bw}.

\subsection{Background: deterministic and stochastic samplers}

A flow matching model is defined according to an ordinary differential equation (ODE), specified by a vector field $\v_t(\x)$ satisfying $d\x_t = \v_t(\x_t) dt$, with initial condition (IC) $\x_0\sim\N(\0, \I)$ at $t=0$.
A flow matching model learns a neural network representation, $\v_t^\theta(\x)$, that minimizes the squared error to a target vector field, $\L(\theta) = \left\Vert \v^\theta_t(\x_t) - \v^{\mathrm{target}}_t(\x_t) \right\Vert^2$,
and in practice, including for VLAs, it is common to define $\v^{\mathrm{target}}_t(\x_t) = \x_1 - \x_0$.
The action $\x_1 = \x_0 + \int_0^1 \v^\theta_t(\x_t) dt$ is computed by integrating the ODE. We perform this integration using the Euler method over $1/\Delta t$ time steps, iteratively computing
\begin{eqnarray}
    \x_{t+\Delta t} = \x_t + \v^\theta_t(\x_t) \Delta t
    \label{eqn:euler}
\end{eqnarray}
to transform a sample of unit Gaussian noise, $\x_0 \sim \N(\0, \I)$, into a robot action $\x_1 \sim \pdata$.
In $\pi_{0.5}$, $\Delta t = 0.1$, meaning the Euler integration takes $10$ steps for inference.

\input{fig2}

In contrast, a diffusion model \cite{Ho2020-lb} is commonly sampled by integrating a stochastic differential equation (SDE), $d\x_t = \v_t(\x_t) dt + \sigma_t d\W_t$, where $\W_t$ is a stochastic Wiener process.
In practice, this SDE can be integrated using the Euler--Maruyama method, which iteratively computes
\begin{eqnarray}
    \x_{t+\Delta t} = \x_t + \v^\theta_t(\x_t) \Delta t + \sqrt{\Delta t} \sigma_t \epsilon_t,
\end{eqnarray}
where $\epsilon_t \sim \N(\0, \I)$, so that noise is added at each time step (Figure~\ref{fig:fig2}a).
A diffusion model can also be sampled deterministically, by integrating its probability flow ODE \cite{Song2021-sde} or with the deterministic sampler of denoising diffusion implicit models (DDIM) \cite{Song2020-jh}.
Conversely, a flow matching model can be sampled stochastically.
We therefore contrast \textit{stochastic} samplers, which inject noise at every step from $\x_0$ to $\x_1$, with \textit{deterministic} samplers, which take a noise-free path (Figure~\ref{fig:fig2}b, c).

\subsection{Initial condition information under stochastic and deterministic sampling}
\label{subsec:ic-information}

\textbf{Stochastic samplers lose initial condition information.}
A stochastic sampler forms a Markov chain
$\x_0 \to \x_{\Delta t} \to \cdots \to \x_1$ given the conditioning $\c$ (camera, state,
language), with each step adding independent noise.
Applying the data processing inequality \citep{Thomas-M-Cover2012-ap} to the sub-chain
$\x_{k\Delta t} \to \x_{(k+1)\Delta t} \to \cdots \to \x_1$ yields, for every $k$,
\begin{align}
    I\!\left(\x_{k \Delta t}; \x_1 \mid \c\right)
    \;<\;
    I\!\left(\x_{(k+1) \Delta t}; \x_1 \mid \c\right),
    \label{eq:dpi-diffusion}
\end{align}
with the inequality being strict because the injected noise is non-degenerate.
The mutual information with the final action is therefore \emph{smallest at $\x_0$} and grows
monotonically toward $\x_1$.
Intuitively, this means that information in the IC $\x_0$ is progressively washed out.
As \citet{yoneda2023noise} observe, a user action injected near $\x_0$ therefore results in a low-fidelity action (not reflecting user input).

\textbf{Deterministic samplers preserve initial condition information.}
We assume the learned velocity field $\v^\theta_t(\cdot\mid\c):\R^d\times[0,1]\to\R^d$ is continuously differentiable and $L$-Lipschitz in $\x$, uniformly in $t\in[0,1]$.
This assumption holds for the smooth networks used in practice, and results in a flow $\psi_t$, mapping $\x_0$ to $\x_t$, that is a diffeomorphism for every $t$ \citep{Lipman2022-oa, Holderrieth2025-hb}.
Consequently, $\x_t$ and $\x_1$ are related, given $\c$, by the deterministic bijection
$\x_t = \psi_t\!\left(\psi_1^{-1}(\x_1)\right)$; in particular the IC is fully
recoverable, $\x_0 = \psi_1^{-1}(\x_1)$.
The data processing inequality therefore holds with equality,
\begin{align}
    I\!\left(\x_{k \Delta t}; \x_1 \mid \c\right)
    \;=\;
    I\!\left(\x_{(k+1) \Delta t}; \x_1 \mid \c\right)
    \qquad \text{for all } k,
    \label{eq:dpi-flow}
\end{align}
in contrast to the strict inequality of \eqref{eq:dpi-diffusion}.
A deterministic sampler therefore \textit{preserves} information about the IC $\x_0$ in the eventual action $\x_1$.
(We note that for continuous variables related by a deterministic bijection, the mutual information in \eqref{eq:dpi-flow} is unbounded, so we use this equality only to express that $\x_0$ is exactly recoverable from $\x_1$, which is not the case for a stochastic sampler.)

Beyond preserving information, we hypothesize that the determinism of the flow makes the initial condition a usable control input.
Because the flow integrates a bounded velocity field, it transports a point only a limited distance: setting $\x_0=\x_\user$, the generated action $\x_1=\psi_1(\x_\user)$ satisfies $\|\x_1-\x_\user\| \le \int_0^1 \|\v^\theta_t(\x_t)\|\,dt$, so $\x_1$ stays relatively close to the injected user action $\x_\user$ (high fidelity to the user input).
At the same time, the flow is trained to transport samples of $\N(\0,\I)$ to $\pdata$, so we expect $\x_1$ to stay close to $\pdata$ when the injected IC has high density under $\N(\0,\I)$, which is typically the case for our scaled inputs (median magnitude $0.46\sigma$, with $96\%$ within $2\sigma$; Appendix~\ref{app:ic-stats}).
We quantify how close actions steered by flow control stay to the action expert distribution in Appendix~\ref{app:onpolicy}.
We further intuit that when there are two possible action modes, initial conditions that are closer to a particular mode are more likely to converge to it (Figure~\ref{fig:fig2}c, green trajectory).
This is because, although flow ODEs do not typically have stable fixed points \textit{per se}, they transport initial conditions toward high-density regions of $p_{\text{expert}}$, which therefore act like fixed points with basins of attraction (Appendix~\ref{app:steering}).
In contrast, with a stochastic sampler, the noise added at every time step can carry the IC along a noisy path toward another mode (Figure~\ref{fig:fig2}c, orange trajectory).
That deterministic samplers preserve information about the initial condition motivates \textit{modifying the initial condition} of the flow model for user control of VLAs.

Using LIBERO simulations \cite{Liu2023-libero}, we trained two small matched policies, one with a flow matching loss and one with a diffusion loss.
We found that IC steerability decreased as the sampler injected more noise for both policies.
When both policies were sampled deterministically, the flow matching policy responded more precisely to IC steering than the diffusion policy (Appendix~\ref{app:samplers}).

\subsection{Our contribution: flow control of VLAs}
\label{subsec:flowcontrol}

\begin{wrapfigure}{r}{0.47\textwidth}
\vspace{-0.35cm}
\footnotesize
\setlength{\parindent}{0pt}\setlength{\parskip}{0pt}
\hrule height 0.9pt
\vspace{2pt}
\refstepcounter{algorithm}\label{alg:flowcontrol}%
{\textbf{Algorithm \thealgorithm:} Flow control\par}
\vspace{1pt}
\hrule height 0.4pt
\vspace{1pt}
\begin{algorithmic}[1]
\Require VLA with flow matching expert $\v^\theta_t$; ODE steps $K$; steered steps $\tau$
\Loop
  \State observe $\c$ (camera, language, state)
  \State $\v_\user \gets$ user keyboard input
  \State $\x_\user \gets \mathrm{IK}(\v_\user)$
  \State sample $\x_0 \sim \N(\0,\I)$
  \State if a key is held: $\x_0[0{:}\tau] \gets \textrm{normalize}(\x_\user)$
  \For{$k = 0,\dots,K-1$}
    \State $\x_{k+1} \gets \x_{k} + \Delta t\,\v^\theta_{t_k}(\x_k\!\mid\!\c)$
  \EndFor
  \State execute the action chunk on the robot
\EndLoop
\end{algorithmic}
\vspace{1pt}
\hrule height 0.9pt
\vspace{-\intextsep}
\end{wrapfigure}

We propose a simple way to achieve real-time user control of VLAs:
we modify the initial condition of the flow ODE with a user input. %
In this work, the user input $\u$ comes from a keyboard, where each held key commands one of six directions: up, down, left, right, forward, or backward.
These inputs are translated into a 3D Cartesian end effector velocity of approximately $11$~cm/s in the direction of the pressed key, $\v_\textrm{user}$.
We then use the regularized inverse kinematics (IK) of the DROID controller \cite{Khazatsky2024-bw} to compute $7$ robot joint velocity commands, $\x_\textrm{user} = \textrm{IK}(\v_\textrm{user})$.
For the command to influence the flow, it must be comparable in scale to the samples of $\N(\0,\I)$ that the action expert was trained on, without being so large that it is improbable under this distribution.
Our joint velocities are small relative to this scale (median norm $0.12$), so our $\textrm{normalize}(\cdot)$ multiplies them by a fixed gain $g = 15$.
We write the result into the $7$ joint-velocity dimensions of the first $\tau$ of $16$ time steps of $\x_0$.
We choose $\tau$ empirically in Section~\ref{subsec:steer}.
We execute $8$ actions before the VLA replans.
All other coordinates of $\x_0$, including the gripper dimension, remain samples of $\N(\0,\I)$, so flow control never commands the gripper.
When no key is held, $\x_0$ is unmodified and the VLA acts autonomously.
In general, $\x_\textrm{user}$ is OOD with respect to the VLA's action expert distribution, $\pdata$.
Motivated by Section~\ref{subsec:ic-information}, the flow matching action expert then iteratively transforms this IC into an action chunk $\x_1$ that still contains information about user inputs while staying close to the action expert distribution (Algorithm~\ref{alg:flowcontrol}, Appendix~\ref{app:impl}).
The command need not come from a person: any source that outputs a direction, such as a scripted controller or a VLM, can write the IC (Appendix~\ref{app:vlm}).
We emphasize that flow control (1) requires no additional training of a VLA, (2) provides a simple and intuitive way to steer a VLA in real time, (3) adds minimal inference latency, and (4) generates actions that stay close to the VLA's own action distribution (Section~\ref{subsec:robust}).

\subsection{Embodiment, tasks, and user study}

\textbf{Embodiment}: All tasks were performed using a Franka Panda robot arm, with one scene camera (ZED) and one wrist camera (ZED Mini) as in DROID \cite{Khazatsky2024-bw}.
We used $\pi_{0.5}$-DROID because it is a state-of-the-art VLA with a flow matching action expert \cite{Intelligence2025-ht}, although flow control can be used with other flow matching policies.
At each inference step, $\pi_{0.5}$-DROID produces an action chunk of $16$ time steps, of which only the first $8$ are executed.
Each action is $8$-dimensional, comprising the $7$ joint velocities of the arm and the gripper position.
The control loop runs at $15$~Hz, so the VLA replans every $0.53$~s.
All inference was performed on a single GPU (NVIDIA GeForce RTX 5090).

\textbf{Tasks}:
We performed four tasks (full descriptions in Appendix~\ref{app:tasks}).
\begin{enumerate}[nosep]
    \item \textbf{Two-Block pick-and-place} (Sections~\ref{subsec:steer} and~\ref{subsec:robust}, Figure~\ref{fig:fig3}a):
    Two blocks are placed on a table with a square hole.
    The policy receives the ambiguous command ``put the block in the hole.''
    A programmed steering command influences which block is picked.
    This is our simplest task, and for interpretability we performed flow control only on joint 1.
    All other tasks used flow control on all $7$ arm joints.
    \item \textbf{Five-Block pick-and-place} (Appendix~\ref{app:ood}, Figure~\ref{fig:fig5}a):
    Five block goals are positioned at varying proximity.
    The policy receives the ambiguous command ``put the block in the hole.''
    This task characterizes the resolution of flow control, and methods to improve it by allowing partially OOD actions.
    \item \textbf{Marker-in-Bowl} (Section~\ref{subsec:userstudy}, Figure~\ref{fig:fig8}a):
    Three differently colored markers are spaced $10$~cm apart in front of a bowl.
    The unambiguous command is ``put the red marker in the bowl.''
    \item \textbf{Cup-Stacking} (Sections~\ref{subsec:userstudy} and~\ref{subsec:finetune}, Figure~\ref{fig:fig8}b):
    Three cups are placed on a table, and the policy must stack them in any order.
    The unambiguous command is ``stack the cups.''
\end{enumerate}

\textbf{User study}:
All experiments were approved by the UCLA IRB.
We performed a user study with $16$ users (Appendix~\ref{app:user}).
Each user performed $10$ trials each of Marker-in-Bowl and Cup-Stacking, providing keyboard inputs to steer $\pi_{0.5}$-DROID, after at most $5$~optional practice trials.
After filtering for valid trials, we had $314$ total trials (Appendix~\ref{app:interventions}).
Reported success rates pool trials across users.
Users also teleoperated the robot arm with the Meta Quest 2 of the DROID system \cite{Khazatsky2024-bw} for an additional $10$ trials of each task.

%% file: fig2.tex
\begin{wrapfigure}{r}{0.5\textwidth}
    \vspace{-0.55cm}
    \centering
    \includegraphics[width=\linewidth]{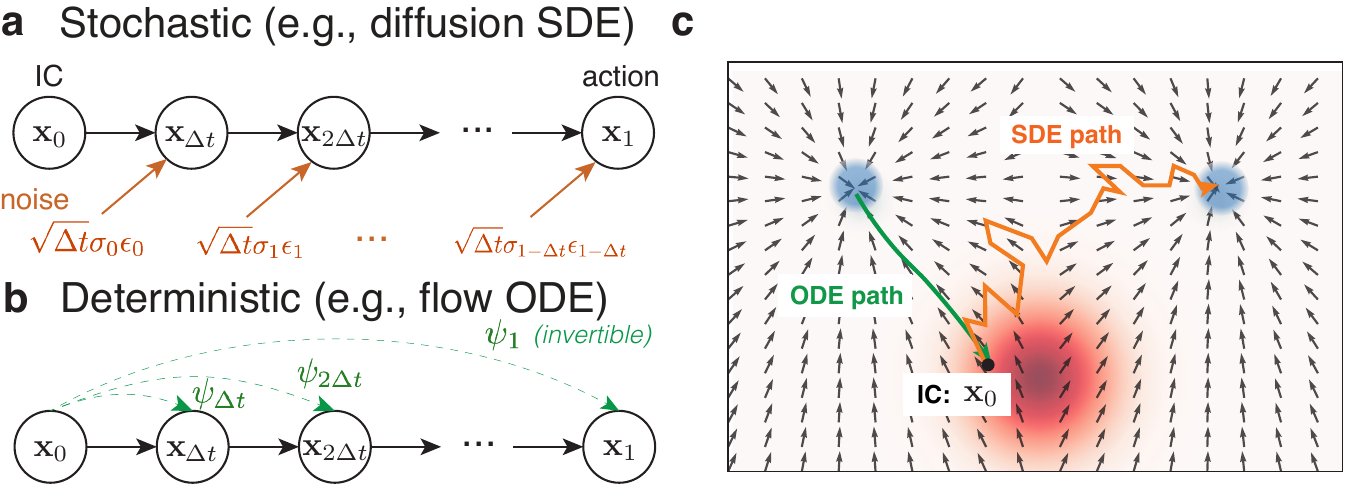}
    \caption{
    \textbf{a}, A stochastic sampler, such as the diffusion SDE, injects noise at each time step, and therefore information in the initial condition (IC) is lost over successive iterations.
    \textbf{b}, A deterministic sampler, such as the flow matching ODE, defines a diffeomorphism and is therefore invertible, preserving information about the IC in $\x_1$.
    \textbf{c},
    The deterministic trajectory integrates an ODE directly to the left mode; the stochastic trajectory integrates a noisy SDE that can arrive at the right mode.
    }
    \label{fig:fig2}
\end{wrapfigure}

%% file: 4_experiments.tex
\section{Experiments}
\label{sec:experiments}

We first empirically assess if injecting user intent through the flow IC is sufficient to steer the VLA policy in a simple setting where flow control is applied to only one joint.
This enables us to also analyze whether steered actions stay close to the VLA's action distribution.
We subsequently perform a user study to show that when users steer $\pi_{0.5}$-DROID with flow control, performance significantly increases.
We further demonstrate that fine-tuning $\pi_{0.5}$ with flow control trajectories increases autonomous performance.

\input{fig3}

\subsection{VLAs can be steered by modifying the flow IC}
\label{subsec:steer}

Using the Two-Block task (Figure~\ref{fig:fig3}a), we provided $\pi_{0.5}$ the ambiguous language instruction ``put the block in the hole.''
$\pi_{0.5}$ autonomously picked the right block $85\%$ of the time.
Our goal was to programmatically inject flow control ICs and determine under what settings, if any, modifying the IC could steer the autonomous policy to pick up the left block instead of the right block.
For interpretability, we made this steering intentionally simple: since the blocks are positioned on the left-to-right axis, we only perturbed joint 1 of the Franka arm (controlling left-to-right movement, Figure~\ref{fig:fig3}b).
To remove any user variability, we applied this steering signal \textit{at every time step}.
Please note that all other experiments performed flow control on all robot arm joints.

To steer $\pi_{0.5}$ toward the left block, we perturbed the joint 1 IC via $\x_0^{[\textrm{joint 1}]} \leftarrow \x_0^{[\textrm{joint 1}]} - \sigma$ with $\sigma=1$ (``add'') or $\x_0^{[\textrm{joint 1}]} \leftarrow - \sigma$ (``set'').
However, because $\pi_{0.5}$ generates an action chunk, there are initial conditions at each of the $T=16$ time steps of the action chunk (Figure~\ref{fig:fig3}c).
We systematically varied the number of time steps of the action chunk with a perturbed IC, $\tau \in \{2, 4, 6, 8\}$.
As $\tau$ increased, we saw the steering command was more effective, with the policy eventually acquiring the left block on nearly $100\%$ of trials (``set'', $\tau=6$ and $\tau=8$, Figure~\ref{fig:fig3}d; we use ``set'' with $\tau=8$ for the rest of the paper).
Finally, we observed that steering the VLA directionally did not adversely affect the VLA's other actions, including picking and placing blocks (Figure~\ref{fig:fig3}e).
Setting the IC of the flow matching action expert therefore predictably steered the VLA toward an intended goal.

\subsection{Flow control actions stay close to the VLA's action distribution}
\label{subsec:robust}

\input{fig4}

We emphasize that for the Two-Block steering experiment, we applied the joint-1 perturbation at every time step, yet perturbed trials still completed the pick, transport, and place subtasks at nearly $100\%$ success (Figure~\ref{fig:fig3}e).
After the policy picks up the left block, the task is unambiguous: the arm must move to the right to place the block in the hole.
The policy's action expert distribution can therefore be thought of as unimodal during these unambiguous subtasks, specifying a single trajectory.
When the arm moved the block toward the hole, the flow control steering signal was still being applied, and, critically, was \textit{discordant} with the VLA's own action.
Even so, the flow matching head transformed these discordant steering signals into actions that did not detrimentally affect task completion.

Flow control steering can therefore be thought of as injecting user intent when action distributions are multimodal, and there are multiple potential options (which block to pick, toy illustration in Figure~\ref{fig:fig4}a).
However, when the action distribution was unimodal, flow control did not interfere with task completion.
One intuition is that the flow IC is driven by a strong flow toward a unimodal distribution (illustration in Figure~\ref{fig:fig4}b, c).
Consistent with this, when we empirically measured the effect of the flow control IC perturbation on the VLA's actions, we found the flow IC was influential before the policy picked one of the two blocks, but not so when flow control steering would impede unambiguous task completion.
This is illustrated in Figure~\ref{fig:fig4}d, where we plot $10$ trajectories of flow control steering with $\tau=8$.
Red in the trajectory indicates high alignment of the VLA action with the flow IC, while blue indicates that the VLA action was discordant with the flow IC.

We therefore emphasize that perturbing the flow IC is not simply adding a joint bias for all movements.
When flow control steering injected user inputs that would negatively impact task completion, the flow matching action expert transformed discordant user input into actions that still completed the task.
In other words, in this experiment the flow IC only steered the policy when it mattered.
This raises a secondary question: what if the user \textit{wanted} to steer the policy away from the VLA's unimodal action expert distribution?
We found it is possible to generate partially OOD actions by also injecting the user action into the ODE integration, which we detail with a Five-Block pick-and-place task in Appendix~\ref{app:ood}.

We also quantified the conformity of flow control actions to the VLA's own actions (Appendix~\ref{app:onpolicy}).
In the LIBERO simulation benchmark \cite{Liu2023-libero} (Appendix~\ref{app:libero}) with $\pi_{0.5}$-LIBERO, we found that approximately $90\%$ of flow control action chunks were within the range of the autonomous VLA's chunks.
Specifically, $88$--$91\%$ had a behavior cloning (BC) loss between the 5th and 95th percentiles of the autonomous chunks, compared to $4$--$14\%$ for adding the user command to the autonomous chunk (action blending) and $0$--$2\%$ for the raw user command (Figure~\ref{fig:bcloss}, Table~\ref{tab:sim-bcloss}).
On the real robot, the median nearest-neighbor distance from flow control chunks to the chunks of autonomous $\pi_{0.5}$-DROID rollouts was $1.08$--$1.36\times$ that of the VLA's own unsteered chunks (Appendix~\ref{app:onpolicy}).
We also tested language corrections, which keep actions on the VLA's distribution by construction, and found that the corrections we tested did not steer the VLA (Appendix~\ref{app:onpolicy}).

\vspace{-0.3cm}

\subsection{Flow control steering significantly increases VLA performance}
\label{subsec:userstudy}

We next performed a user study to assess if flow control could steer VLAs to achieve better task performance.
We tested two tasks with $n=16$ participants.
In the Marker-in-Bowl task, where the VLA was unambiguously prompted to ``put the red marker in the bowl'' (Figure~\ref{fig:fig8}a), $\pi_{0.5}$-DROID demonstrated poor language following: while it always placed a marker in the bowl, it was the correct marker on only $53.3\%$ of trials.
In this case, users steered the policy with keyboard inputs to correct incorrect language following, increasing the success rate of picking the correct marker to $99.4\%$ (Table~\ref{tab:auto_vs_flow}, $p<0.05$, Wilcoxon rank sum).
In this task, both autonomous and flow control execution were fast, with flow control being $8.3\%$ faster ($p<0.05$, Wilcoxon rank sum).
Flow control steering therefore enabled users to overcome incorrect language following, substantially increasing task success rate.

\begin{wraptable}{r}{0.35\textwidth}
  \centering
  \vspace{-\intextsep}
  \footnotesize
  \setlength{\tabcolsep}{4pt}
  \begin{tabular}{lcc}
    \toprule
     & \textbf{Autonomous} & \textbf{FC} \\
    \midrule
    \multicolumn{3}{l}{\textit{Success rate} ($\uparrow$)} \\
    \quad Marker & 53.3\% & \textbf{99.4\%} \\
    \quad Stack  & 48.0\% & \textbf{87.7\%} \\
    \midrule
    \multicolumn{3}{l}{\textit{Trial time} [s] ($\downarrow$)} \\
    \quad Marker & 14.50 & \textbf{13.30} \\
    \quad Stack  & 50.33 & \textbf{35.21} \\
    \bottomrule
  \end{tabular}
  \caption{Performance of autonomous vs flow control (FC) steering.
  }
  \label{tab:auto_vs_flow}
  \vspace{-\intextsep}
\end{wraptable}

In the Cup-Stacking task, the VLA was prompted to stack three cups (Figure~\ref{fig:fig8}b).
This task was relatively long horizon and more challenging than prior tasks.
We found $\pi_{0.5}$-DROID was able to perform this task successfully $48.0\%$ of the time, with a common failure mode being that the policy knocked over cups while trying to stack them (Figure~\ref{fig:fig8}c).
On successful trials, $\pi_{0.5}$-DROID's average trial time was $50.33$~seconds.
When users steered the policy with flow control, the success rate increased significantly to $87.7\%$ ($p<0.05$, Wilcoxon rank sum).
Users frequently corrected the policy, for example, when the robot arm tried to stack the cups without lifting a held cup high enough (Figure~\ref{fig:fig8}d).
Overall, users held a key on $17.9\%$ of time steps (per-user range $8.3$--$30.3\%$; Appendix~\ref{app:interventions}).
When a key was held, the first generated action had a median cosine similarity of $0.76$ with the command, and the similarity was positive for $98.0\%$ of steered chunks.
With flow control steering, users significantly decreased the average trial time to $35.21$~seconds, which was $30\%$ faster ($p<0.05$, Wilcoxon rank sum, Table~\ref{tab:auto_vs_flow} and Figure~\ref{fig:fig8}e).

\input{fig8}

We also assessed if flow control steering could help some users perform the task faster than teleoperation, which can be challenging to master (Appendix~\ref{app:teleop}).
In summary, we found that the across-user variance in completion time was $32\times$ ($2.8 \times$) smaller for Marker-in-Bowl (Cup-Stacking) with flow control than with teleoperation.
Flow control therefore equalized performance across operators, with na\"ive teleoperators performing the Marker-in-Bowl (Cup-Stacking) task $50.3\%$ ($14.7\%$) faster with flow control (Appendix~\ref{app:teleop}).
Finally, in LIBERO, where a scripted controller provided the inputs, flow control raised the task success of $\pi_{0.5}$-LIBERO from $27/50$ to $47/50$ and from $23/50$ to $33/50$ on two tasks (Appendix~\ref{app:libero}).

\subsection{Fine-tuning $\pi_{0.5}$ on flow control steering trajectories improves VLA performance}
\label{subsec:finetune}

Flow control actions stay close to the VLA action distribution (Section~\ref{subsec:robust}), and therefore provide action trajectories -- with corrections -- that can be used to improve the policy.
We fine-tuned $\pi_{0.5}$-DROID with $60$ successful flow control trials from a separate session, using standard behavior cloning on the executed actions (keypresses were not used; details in Appendix~\ref{app:finetune}); we call the resulting policy $\pi_{0.5}$-FC.
Autonomous $\pi_{0.5}$-FC achieved a $100\%$ success rate on the Cup-Stacking task over $25$ trials (compared to $48\%$ for $\pi_{0.5}$-DROID), and decreased the median trial time to $22$~seconds (Figure~\ref{fig:fig8}e).
Flow control therefore not only provides the ability for users to steer VLA trajectories in real time, but also yields data that durably improves the underlying policy.
Because flow control trajectories are also collected with simple inputs rather than full teleoperation, reducing the across-user variance in completion time (Appendix~\ref{app:teleop}), they may be cheaper to gather at scale.
Finally, in simulation with LIBERO, we found that fine-tuning $\pi_{0.5}$-LIBERO on flow control successes raised its autonomous success from $41.1\%$ to $75.6\%$, whereas fine-tuning on its own autonomous successes reached $47.8\%$ (Appendix~\ref{app:finetune}, Figure~\ref{fig:simft}).

%% file: fig3.tex
\begin{figure*}
    \centering
    \includegraphics[width=\linewidth]{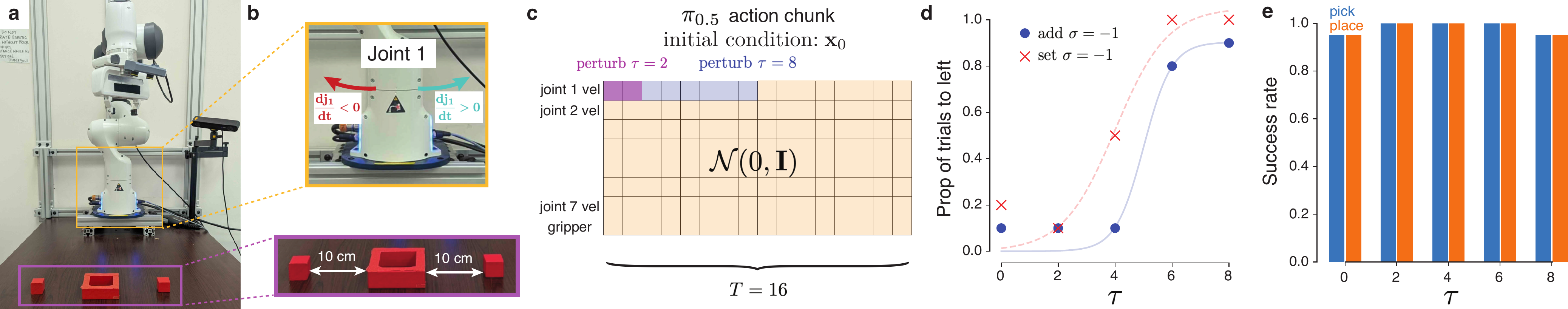}
    \caption{
    \textbf{a}, Task, ``put the block in the hole.''
    The $2$~cm blocks are each $10$~cm from the hole.
    \textbf{b}, Joint 1 controls left-right movement.
    \textbf{c}, We perturb joint 1's initial condition between $\tau=0$ and $\tau=8$ time steps over the horizon of $16$ time steps in the $\pi_{0.5}$ action chunk.
    \textbf{d}, As $\tau$ increases, the proportion of left trials increases.
    \textbf{e}, Perturbing the IC does not adversely affect pick or place performance on the task.
    }
    \label{fig:fig3}
    \vspace{-0.5cm}
\end{figure*}

%% file: fig4.tex
\begin{wrapfigure}{r}{0.5\textwidth}
    \centering
    \vspace{-\intextsep}
    \includegraphics[width=\linewidth]{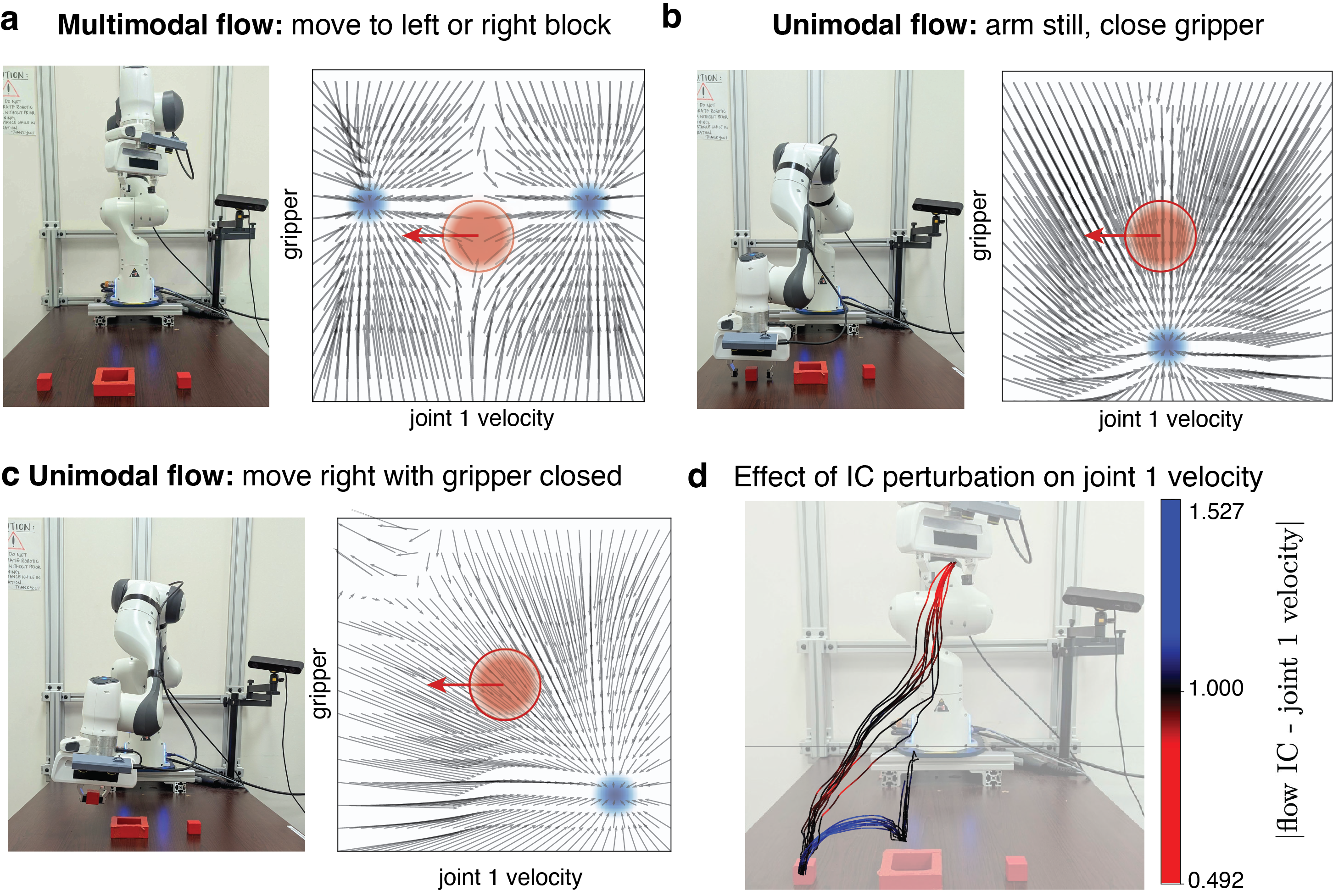}
    \caption{
    \textbf{a}, In a multimodal setting, flow control steers the policy toward the left block.
    \textbf{b, c}, In a unimodal setting, the flow IC is transformed into an action that still completes the task.
    \textbf{d}, In experiments, the IC perturbation only affects the trajectory early (red) and not when the task is unambiguous (black and blue).
    }
    \label{fig:fig4}
    \vspace{-\intextsep}
\end{wrapfigure}

%% file: fig8.tex
\begin{figure}[t!]
    \centering
    \includegraphics[width=\linewidth]{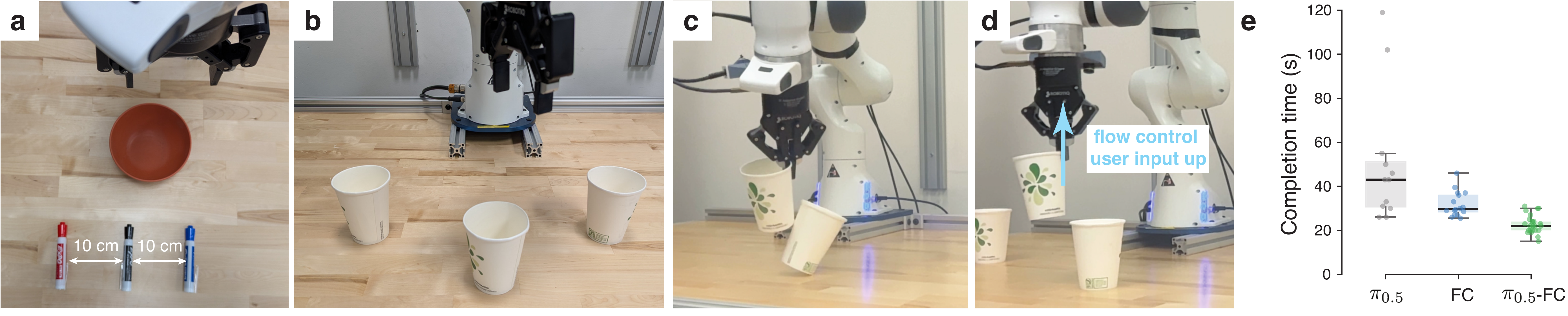}
    \caption{
    \textbf{a}, Marker-in-Bowl task.
    \textbf{b}, Cup-Stacking task.
    \textbf{c}, Example failure of $\pi_{0.5}$-DROID, not lifting a cup high enough and knocking over another.
    \textbf{d}, The user gives a flow control input ``up'' to avoid this failure.
    \textbf{e}, Trial time distribution for Cup-Stacking for the autonomous, flow control, and flow control fine-tuned VLA.
    }
    \label{fig:fig8}
    \vspace{-0.6cm}
\end{figure}

%% file: 5_discussion.tex
\section{Discussion}

We demonstrate that, after setting the initial condition of the flow matching action expert to be the user action, it is possible for users to steer VLAs to improve task success and speed.
This flow control steering can be used zero-shot with policies that use flow matching, including VLAs with a flow matching action expert, and does not require any fine-tuning.
Further, we demonstrate that flow control actions stay close to the actions the VLA generates on its own, and that these trajectories can further fine-tune VLAs to improve autonomous performance.
Flow control also enables na\"ive teleoperators to achieve significantly faster task completion times, and can therefore be viewed as a form of shared autonomy.
As flow control provides simple, intuitive, and real-time steering of VLAs through keyboard inputs, we anticipate this method may help to correct and adjust robot actions when autonomous policies perform suboptimally.

In our user study, a person provided the steering inputs.
However, these inputs can also come from a VLM.
In a separate set of simulation runs, a VLM controller raised task success from $28/50$ to $46/50$ and from $22/50$ to $34/50$ on two LIBERO tasks, matching a scripted controller (Appendix~\ref{app:vlm}), so agentic robot systems could steer VLAs using flow control.

\subsection{Limitations}

While we believe a benefit of flow control is that it keeps actions close to the VLA's action distribution, this could also be viewed as a limitation.
For example, if the VLA ``commits'' to a certain task goal, user inputs will have little impact on robot actions.
While we partially addressed this by perturbing the flow process in Appendix~\ref{app:ood}, we by no means suggest this is an optimal way to address this limitation.
Flow control also requires a competent policy: it steers among the actions the VLA already generates, so it generally does not produce a behavior that the VLA does not perform on its own.
Future work may therefore investigate how to more effectively sample OOD actions.
A further limitation is that we evaluated only one VLA family, which uses a flow matching action expert, on one physical embodiment (Appendix~\ref{app:impl}).
VLAs with diffusion action experts can be steered through the initial condition when sampled deterministically, although less precisely than flow matching in our tests (Appendix~\ref{app:samplers}).
VLAs with stochastic samplers may adopt steering methods from the shared autonomy literature \cite{yoneda2023noise, Wang2026-ko}.
This method does not apply to VLAs that autoregressively generate discrete action tokens, which can instead be steered through their action tokens \cite{Chan2026-ts}.

%% file: appendix.tex
\newpage
\input{app_geometry}

\newpage
\input{app_mod_samplers}

\newpage
\input{app_tasks}

\newpage
\input{app_control}

\newpage
\input{app_teleop}

\newpage
\input{app_implementation}

\newpage
\input{app_mod_simulation}

%% file: app_geometry.tex
\section{Geometry of initial condition steering}
\label{app:steering}

This appendix expands on the intuition for why modifying the initial condition can steer the flow matching action expert (Section~\ref{subsec:ic-information}).
We emphasize that this appendix provides intuition only, and that the primary evidence for flow control is empirical (Section~\ref{sec:experiments}).

\textbf{Flow displacement.}
Under the Lipschitz assumption on the velocity field $\v^\theta_t(\cdot\mid\c)$ (Section~\ref{subsec:ic-information}), the flow $\psi_t$ is a diffeomorphism: it is invertible and the initial condition is fully recoverable, $\x_0 = \psi_1^{-1}(\x_1)$.
We make two observations for \emph{any} shape of $\pdata$.
First, integrating the ODE bounds how far the flow can transport an initial condition:
\begin{equation}
    \big\| \psi_1(\x_\user) - \x_\user \big\|
    = \left\| \int_0^1 \v^\theta_t(\x_t\mid\c)\,dt \right\|
    \le \int_0^1 \big\| \v^\theta_t(\x_t\mid\c) \big\|\,dt.
    \label{eq:displacement}
\end{equation}
We define the \textit{displacement} as
\begin{eqnarray}
    D(\x_\user) = \int_0^1 \big\| \v^\theta_t(\x_t\mid\c) \big\|\,dt,
\end{eqnarray}
which is intuitively the sum of vector field displacements over ODE integration.
Second, the flow is trained so that $\psi_1(\x_0)\sim\pdata(\cdot\mid\c)$ for $\x_0\sim\N(\0,\I)$, so we expect an $\x_\user$ that has high density under $\N(\0,\I)$ to produce an action close to $\pdata$.
We emphasize it is important to normalize $\x_\user$ to the scale of the prior before injecting it (Appendix~\ref{app:impl}).
By \eqref{eq:displacement}, the generated action differs from the injected $\x_\user$ by at most $D(\x_\user)$.
Ultimately, however, this does not guarantee that an individual $\x_\user$ of high prior density is mapped to an action close to the VLA's own actions, so we measure this empirically (Appendix~\ref{app:onpolicy}).

\textbf{Basins from the invertibility of flow and a geometric intuition.}
In Section~\ref{subsec:ic-information}, we referred to ``basins-of-attraction'' for the flow.
Because $\psi_1$ is a bijection given $\c$, any partition $\{\mathcal{R}_k\}$ of action space induces a partition of initial-condition space into the preimages $\mathcal{B}(\mathcal{R}_k) = \psi_1^{-1}(\mathcal{R}_k)$, and the prior mass of each basin equals the expert mass of its region,
\begin{eqnarray*}
    \int_{\mathcal{B}(\mathcal{R})} \N(\x;\0,\I)\,d\x = \int_{\mathcal{R}} \pdata(\x\mid\c)\,d\x.
\end{eqnarray*}
A basin is therefore a preimage under the flow.
The basins-of-attraction picture additionally requires a \emph{geometric} property that invertibility alone does not provide: that a basin $\mathcal{B}(\mathcal{R})$ lies close to its own region $\mathcal{R}$, so that injecting $\x_\user$ near a desired action lands in the basin that produces it.
If the displacement $D(\x_\user)$ is small relative to the separation between candidate actions, the generated action stays near the injected $\x_\user$ and commits to the behavior the user indicated. Concretely, for any two candidate actions $\a, \a'$ (e.g., reaching left versus right), if
\begin{equation}
    \|\x_\user - \a'\| - \|\x_\user - \a\| \;>\; 2\,D(\x_\user),
    \label{eq:selection}
\end{equation}
then $\x_1 = \psi_1(\x_\user)$ is closer to $\a$, since $\|\x_1 - \a\| \le \|\x_\user - \a\| + D(\x_\user)$ while $\|\x_1 - \a'\| \ge \|\x_\user - \a'\| - D(\x_\user)$.
We stress that \eqref{eq:selection} is only a \emph{conditional}.
Whether the trained flow is well-behaved enough for \eqref{eq:selection} to hold is therefore an empirical matter, which we defer to our experiments (Section~\ref{sec:experiments}).
In our experiments, simple keyboard inputs reliably steered $\pi_{0.5}$ toward the intended behavior and disambiguated nearby targets, which is consistent with the displacement being small relative to the separation between candidate actions.

%% file: app_mod_samplers.tex
\section{Deterministic and stochastic samplers with matched policies}
\label{app:samplers}

We evaluate how much IC steering depends on the sampler being deterministic, and whether it differs between flow matching and diffusion (Section~\ref{subsec:ic-information}).
We trained two small policies that differ only in their training objective (flow matching vs diffusion).
We then sampled each model deterministically and stochastically and measured how much of a command written into the IC aligns with the generated actions.

\textbf{Matched policies.}
We trained a flow matching model and a diffusion model on $238$ of the $432$ demonstration episodes of LIBERO-Spatial \citep{Liu2023-libero}.
This subset has $29{,}339$ frames and covers all $10$ tasks, with $18$--$29$ episodes each.
The inputs were the agent-view and wrist camera images at $224\times224$ and the $8$-D robot state.
The outputs were chunks of $10$ actions $\times$ $7$ dimensions.
We encoded each camera image with its own ResNet-18 \citep{He2016-resnet}.
These policies have no language encoder.
Instead, each of the $10$ tasks has a learned embedding vector that specifies the task.
During training, we withheld this embedding with probability $0.5$, replacing it with a shared learned vector, so that the policy can also act from the images and the robot state alone.
We ran all experiments in this appendix without the task embedding, so the policy was not told which object to pick.
The action model is a DiT-style transformer \citep{Peebles2023-dit} over the $10$ chunk tokens.
The network has $28.35$M parameters and is identical for both the flow matching and diffusion models.
We normalized with per-dimension 1st and 99th percentile statistics, which were shared by both models.
We trained both models with AdamW for $60{,}000$ steps, with batch size $64$ and a peak learning rate of $10^{-4}$.
Both models used seed $0$ and seeded generators for the batch order and the task dropout.
The two models therefore saw identical batches.

The two models differ in their objective.
Let $\eps\sim\N(\0,\I)$ and let $\a$ be an action chunk.
The flow matching model regresses $\v^\theta_t(\x_t)$ onto $\a-\eps$ at $\x_t=(1-t)\,\eps+t\,\a$, where $1-t=0.001+0.999\,u$ and $u\sim\mathrm{Beta}(1.5,1)$ \citep{Lipman2022-oa, Black2024-zr}.
The diffusion model is a denoising diffusion probabilistic model (DDPM) \citep{Ho2020-lb} with $S=100$ steps, a cosine noise schedule, and noise prediction.
Both losses use mean squared error.

\textbf{Samplers.}
We sampled the flow matching policy with Euler steps on the flow ODE, as $\pi_{0.5}$ does.
We also sampled the same trained policy stochastically, by integrating the SDE
\begin{equation}
    d\x_t = \Big[\v^\theta_t(\x_t) + \tfrac{w_t}{2}\,\s_t(\x_t)\Big]dt + \sqrt{w_t}\,d\W_t,
    \qquad
    \s_t(\x) = -\frac{\x - t\,\v^\theta_t(\x)}{1-t},
    \qquad
    w_t = \eta\,(1-t),
    \label{eq:flow-sde}
\end{equation}
where $\s_t$ is the score of the marginal of $\x_t$, which the linear path determines from the velocity.
The score term compensates for the injected noise, so that this SDE has the same marginals as the ODE for every $\eta\ge0$ \citep{Song2021-sde}.
Setting $\eta=0$ gives the ODE, and the injected noise has total variance $\eta/2$, which is equal to that of the prior at $\eta=2$.
We sampled the diffusion policy with DDIM \citep{Song2020-jh}, which is deterministic for $\eta=0$ and partially stochastic for $\eta=0.5$, as well as with DDPM sampling~\cite{Ho2020-lb}.
We note that $\eta$ is defined separately for each policy.
In unsteered experiments, we observed (over the $10$ LIBERO-Spatial tasks with $10$ episodes each) that the flow matching policy with ODE sampling achieved $88$ successes out of $100$ trials while the diffusion policy with DDIM, $\eta=0$, achieved $79$ successes out of $100$ trials.

\textbf{Experiments.}
For every sampler, we drew the IC from $\N(\0,\I)$ and overwrote the first $\tau=5$ of its $10$ rows.
In these rows, we set the commanded translation dimension to $\pm\sigma$ with $\sigma=1.5$, and the other five translation and rotation dimensions to $0$.
We used $64$ observations, which we drew at random from the training frames.
For each observation, we generated one chunk with $+\sigma$ injected and one chunk with $-\sigma$ injected, and held the rest of the IC and the sampler's noise seed fixed between the two.
The \emph{on-axis} response (On) is the signed difference between the two chunks on the commanded dimension, and the \emph{off-axis} response (Off) is the mean absolute difference on the other two translation dimensions.
We report both in units of $10^{-3}$ of the raw LIBERO action, as means over the three translation axes and five pairs of noise seeds.

\textbf{Results.}
We ran every sampler except DDPM with $10$, $20$, $50$, and $100$ steps (Table~\ref{tab:samplers-steps}), and first examined the dependence on sampler noise.
Both deterministic samplers were steerable.
For both policies, steerability decreased as the sampler injected more noise, and fell to zero for DDPM.
With $100$ steps, a stochastic sampler of the \textit{same} flow matching policy also lost most of its response (on-axis response of $2.5$ at $\eta=2$, compared to $59.9$ for the ODE).
In other words, steerability through the IC depended on the noise that the sampler injected, for both policies.

\begin{figure}[t!]
\centering
\includegraphics[width=\linewidth]{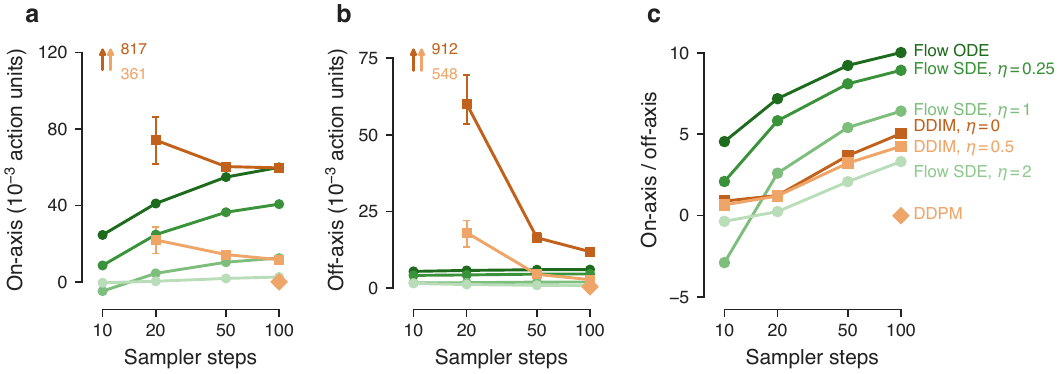}
\caption{The entries of Table~\ref{tab:samplers-steps} as a function of the number of sampler steps.
\textbf{a}, On-axis response. \textbf{b}, Off-axis response. \textbf{c}, Their ratio.
Green denotes samplers of the flow matching policy, and orange denotes samplers of the diffusion policy.
Points are means over the three translation axes and five pairs of noise seeds.
Bars span the five seed pairs, and are smaller than the markers where not visible.
Arrows mark DDIM with $10$ steps, which is off the scale (values next to them).}
\label{fig:sampler-steps}
\end{figure}

We then compared the two deterministic samplers (the flow ODE and DDIM with $\eta=0$) as a function of the number of steps (Figure~\ref{fig:sampler-steps}).
With $100$ steps, they transmitted the command equally ($59.9$ for the flow ODE and $59.6$ for DDIM with $\eta=0$), but DDIM moved the off-axis dimensions twice as much ($11.8$ compared to $6.0$).
With the $10$ steps that $\pi_{0.5}$ uses, the flow ODE still responded cleanly (On $24.5$, Off $5.4$), whereas DDIM generated actions about four times too large.
With $20$ steps, the on-axis response of DDIM varied by a factor of $1.6$ to $2.2$ (depending on the axis) across the five seed pairs, whereas that of the flow ODE varied by at most $1\%$ at every step count.
In other words, both policies were steerable when we sampled them deterministically, but steering the flow matching policy was more precise than steering the diffusion policy.

\begin{table}[h]
\centering
\footnotesize
\setlength{\tabcolsep}{4pt}
\begin{tabular}{@{}llrrrrrrrr@{}}
\toprule
 & & \multicolumn{2}{c}{$10$ steps} & \multicolumn{2}{c}{$20$ steps} & \multicolumn{2}{c}{$50$ steps} & \multicolumn{2}{c}{$100$ steps} \\
\cmidrule(lr){3-4}\cmidrule(lr){5-6}\cmidrule(lr){7-8}\cmidrule(l){9-10}
Model & Sampler & On & Off & On & Off & On & Off & On & Off \\
\midrule
    Flow & ODE (Euler) & $24.5$ & $5.4$ & $41.0$ & $5.7$ & $54.8$ & $5.9$ & $59.9$ & $6.0$ \\
    Flow & SDE, $\eta=0.25$ & $8.6$ & $4.1$ & $24.7$ & $4.2$ & $36.4$ & $4.5$ & $40.6$ & $4.6$ \\
    Flow & SDE, $\eta=1$ & $-4.9$ & $1.7$ & $4.4$ & $1.7$ & $10.3$ & $1.9$ & $12.5$ & $1.9$ \\
    Flow & SDE, $\eta=2$ & $-0.5$ & $1.4$ & $0.3$ & $1.1$ & $1.8$ & $0.8$ & $2.5$ & $0.8$ \\
\midrule
    Diffusion & DDIM, $\eta=0$ & $817$ & $912$ & $74.2$ & $60.1$ & $60.3$ & $16.4$ & $59.6$ & $11.8$ \\
    Diffusion & DDIM, $\eta=0.5$ & $361$ & $548$ & $21.9$ & $18.1$ & $14.3$ & $4.4$ & $11.5$ & $2.7$ \\
    Diffusion & DDPM & -- & -- & -- & -- & -- & -- & $0.0$ & $0.4$ \\
\bottomrule
\end{tabular}
\par\vspace{4pt}
\caption{On-axis (On) and off-axis (Off) response with the same number of steps for every sampler.
Values are in units of $10^{-3}$ of the raw LIBERO action ($\tau=5$, $\sigma=1.5$, $64$ observations).
Each entry is the mean over the three translation axes and five pairs of noise seeds.
DDPM sampling is defined on all $100$ diffusion steps.
With $10$ steps, DDIM generated actions about four times larger than normal.
Its entries in those columns therefore do not measure steering.
$\eta$ sets the injected noise; a sampler with $\eta=0$ is deterministic.}
\label{tab:samplers-steps}
\end{table}

%% file: app_tasks.tex
\section{Task and user study details}

\subsection{Task details}
\label{app:tasks}

All tasks used a Franka Panda arm with a frozen $\pi_{0.5}$ VLA, one scene camera (ZED), and one wrist camera (ZED Mini), following the DROID setup \cite{Khazatsky2024-bw}.

\textbf{Two-Block pick-and-place.}
We placed two $2 \times 2$~cm red blocks at a distance of $10$~cm to the left and right of a square hole.
The square hole had an outer diameter of $6$~cm and an inner diameter of $4$~cm.
The VLA received the ambiguous language command, ``put the block in the hole.''
The goal of the task was to select one block and place it in the hole.
On $85\%$ of trials, $\pi_{0.5}$ selected the right block.
This experiment therefore allowed us to investigate how strongly flow control initial condition perturbations could steer the policy.
This task setup is shown in Figure~\ref{fig:fig3}a, b.

\textbf{Five-Block pick-and-place.}
We placed five $2 \times 2$~cm blocks with different colors (red, black, yellow, blue, and green) in a line (Figure~\ref{fig:fig5}a).
We performed three variants of this task, where the blocks were separated by $0$~cm, $2$~cm, and $4$~cm (Figure~\ref{fig:fig5}a).
The VLA received the ambiguous language command, ``put the block in the hole.''
On every trial, we randomly generated a ``correct'' block color, and steered the VLA with flow control to attempt to acquire the correct block.
This experiment enabled us to assess the precision of flow control, and motivated perturbing the flow process to generate OOD actions (Appendix~\ref{app:ood}).

\textbf{Marker-in-Bowl.}
In this user study experiment, a red, a black, and a blue marker were placed $10$~cm apart in front of a bowl (Figure~\ref{fig:fig8}a).
The markers were lightly taped to the surface to prevent them from rolling.
The VLA received the unambiguous language instruction: ``put the red marker in the bowl.''
If the robot picked up the black or the blue marker, we called the task a failure.
This task tested the language-following ability of $\pi_{0.5}$-DROID.
On $46.7\%$ of trials, the policy incorrectly picked up the black or blue marker and placed it in the bowl.
A trial was also deemed a failure if the marker rolled off the table.

\textbf{Cup-Stacking.}
In this user study experiment, three cups were placed on a table.
We placed the three cups in one of five configurations: in a straight line, along one of two diagonals, or in one of two V shapes.
We did not control the distances between the cups, which varied from trial to trial.
The VLA received the unambiguous language instruction: ``stack the cups.''
The cups could be stacked in any order.
A trial was counted as successful if all three cups were stacked together.
The autonomous policy, flow control, and teleoperation conditions had at most $2$~minutes to complete the task before the trial was deemed a failure.
If a cup fell off the table, the trial was deemed a failure.
A common failure mode in this task was that a cup was knocked over.
We note that $\pi_{0.5}$-DROID was capable of recovering from these states, although this generally increased trial times.
This task was relatively challenging, with $\pi_{0.5}$-DROID successfully completing it $48.0\%$ of the time.

\subsection{User study}
\label{app:user}

Sixteen participants completed both the Marker-in-Bowl and Cup-Stacking tasks under flow control and teleoperation.
They performed $10$ trials per condition, and were given up to $5$ practice trials if they desired.
Teleoperation was performed through VR with the Meta Quest~2 via DROID~\cite{Khazatsky2024-bw}.
All experiments were approved by the UCLA IRB.
Participants ranged from $20$ to $32$ years old; five participants were female and eleven participants were male.
Participants were paid $\$20$ per hour for participating.

%% file: app_control.tex
\section{Control resolution and out-of-distribution steering}
\label{app:ood}

This appendix expands on the results from Section~\ref{subsec:robust} to quantify the precision of flow control steering through IC perturbation, and to show how to steer toward actions that are OOD with respect to the action expert distribution, should these actions be desired.

\subsection{Precision of flow IC control}
\label{app:precision}

\input{fig5}

What precision can flow control achieve in a pick-and-place task?
To answer this question, we performed the Five-Block pick-and-place task at varying difficulties ($0$~cm, $2$~cm, and $4$~cm separation, Figure~\ref{fig:fig5}a).
On each trial, a random block was designated as the correct block to pick.
A user then provided real-time arrow key inputs to guide the VLA to select the correct block out of five; for example, if the VLA was hovering over the yellow block but the correct block was red (Figure~\ref{fig:fig5}a), the user would press the up key to guide the robot arm to the red block.
Occasionally, for the $0$~cm separation condition, the gripper was wide enough to pick and place two blocks at once; in these cases, if one of the two blocks was the prompted block, we counted the trial as successful.

When blocks were separated by $0$~cm, the success rate was $67\%$, with $27\%$ of the trials being off by one block.
This success rate increased to $87\%$ when blocks were separated by $2$~cm (all errors off by one block) and $100\%$ when blocks were separated by $4$~cm.
These results demonstrate that as objects become closer together, accurate steering may occasionally select the wrong goal.
We primarily observed that when the VLA ``committed'' to a block to grasp, its action distribution appeared unimodal, generating trajectories to pick that block: flow control inputs did not steer the policy.
In other words, in these states, modifying the flow IC struggled to ``change the VLA's mind.''

\subsection{Perturbation of the flow to escape unimodal behavior}

\input{fig6}

The experiments in Appendix~\ref{app:precision} led us to ask: can perturbation of the flow ODE integration induce ``corrective'' or ``change of mind'' behavior in the VLA?
To assess this, we returned to the Two-Block task (Figure~\ref{fig:fig3}a) and programmed a perturbation of the flow process to activate at different fixed time intervals in the trial: $1$~s, $2$~s, and $3$~s after the first action taken in the environment (Figure~\ref{fig:fig6}a).
At later times in the trial, the VLA is more strongly committed to picking the right block, and therefore requires stronger perturbations to cause it to move toward the left block.

To counteract this, we not only perturbed the flow IC by $\x_0^{[\textrm{joint 1}]} \leftarrow - \sigma$, but also perturbed the flow for a varying number of steps.
For $\pi_{0.5}$, $\Delta t=0.1$ for the action expert, meaning that integrating the flow ODE computes $\x_0, \x_{0.1}, \x_{0.2}, \ldots, \x_{0.9}, \x_1$.
In these experiments, we also perturbed the ODE integration for $0\leq M \leq 10$ time steps from $t=0$ to $t=M\Delta t$, each time performing $\x_{k \Delta t}^{[\textrm{joint 1}]} \leftarrow \x_{k \Delta t}^{[\textrm{joint 1}]} - \frac{\sigma}{\sqrt{1/\Delta t}}$ after the Euler step, for $k=1, \dots, M$.

We swept $M$ and measured its effect on being able to cause the arm to move toward the left block.
When the perturbation was applied $1$~s into the trajectory, we observed $60\%$ of trajectories moved toward the left goal through only IC control.
Further, when we set $M=4$, so that we perturbed the first four (out of ten) flow steps, the robotic arm always moved in the direction of the left goal.
When perturbations were applied $2$~s into the trajectory, we required $M=6$ to cause the trajectory to change to the left block on every trial.
Finally, when perturbations were applied $3$~s into the trajectory, we required $M=8$ to cause the trajectory to change to the left block, meaning that we had to perform perturbations late into the flow ODE.
Together, these results are intuitive: they demonstrate that perturbing the flow ODE toward desired actions can ``change the VLA's mind'' by putting it in new states the unperturbed VLA action expert would not have converged to.
We strongly caveat, however, that these actions may be OOD and have low probability under $\pdata$.

We subsequently repeated the Five-Block task with the $0$~cm separation condition (i.e., all blocks touching), enabling user inputs to perturb the flow up to $M=7$ ODE steps.
When we perturbed not only the flow IC, but also the flow process, the VLA with flow control selected the correct block on $15$ out of $15$ trials where a block color was randomly prompted.
This indicates that perturbation of the flow process can provide increased control, including movement toward a goal separate from that determined by the VLA.

%% file: fig5.tex
\begin{wrapfigure}{r}{0.5\textwidth}
    \centering
    \vspace{-\intextsep}
    \includegraphics[width=\linewidth]{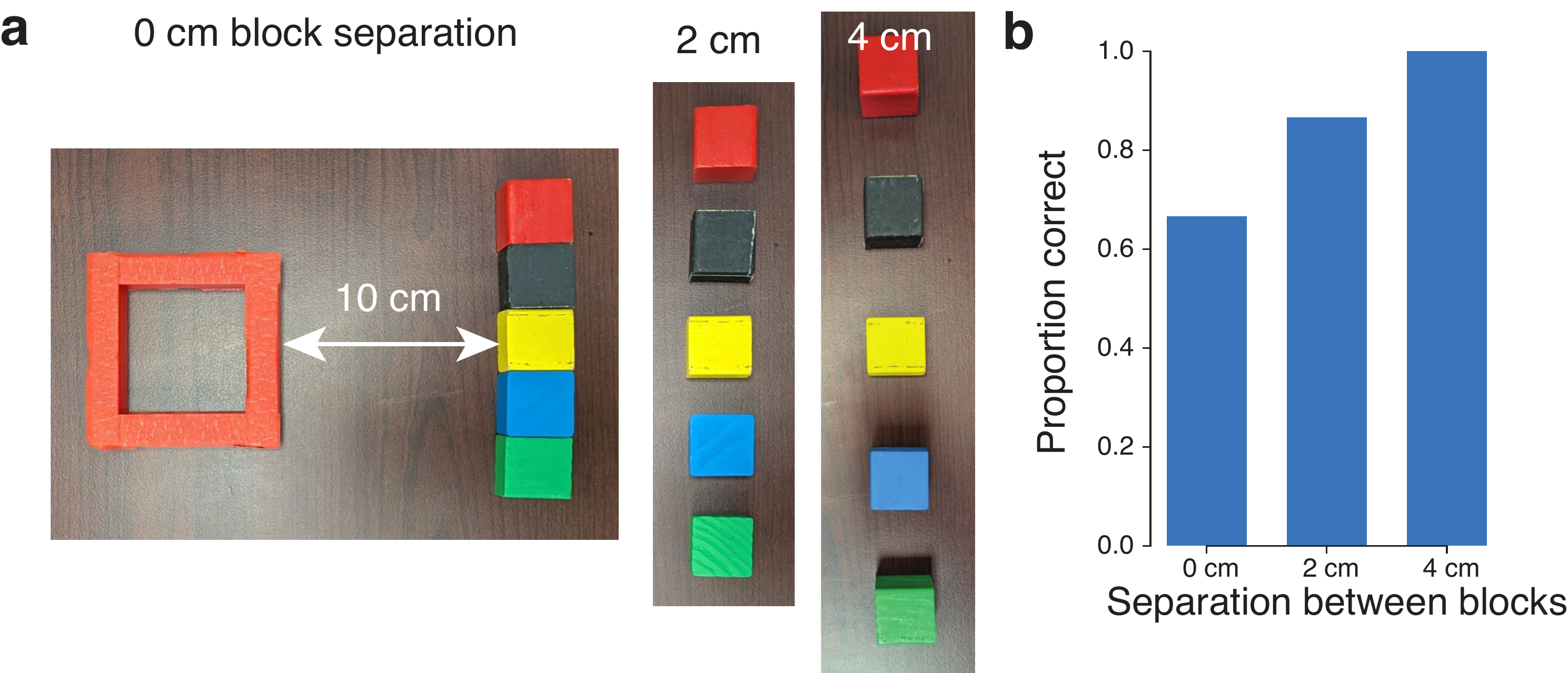}
    \caption{Five-Block task.
    \textbf{a}, Task where blocks are separated by $0$~cm, $2$~cm, or $4$~cm.
    A correct block is randomly specified on each trial.
    \textbf{b}, Success rate of picking up the correct block.
    }
    \label{fig:fig5}
    \vspace{-\intextsep}
\end{wrapfigure}

%% file: fig6.tex
\begin{wrapfigure}{r}{0.5\textwidth}
    \centering
    \vspace{-\intextsep}
    \includegraphics[width=\linewidth]{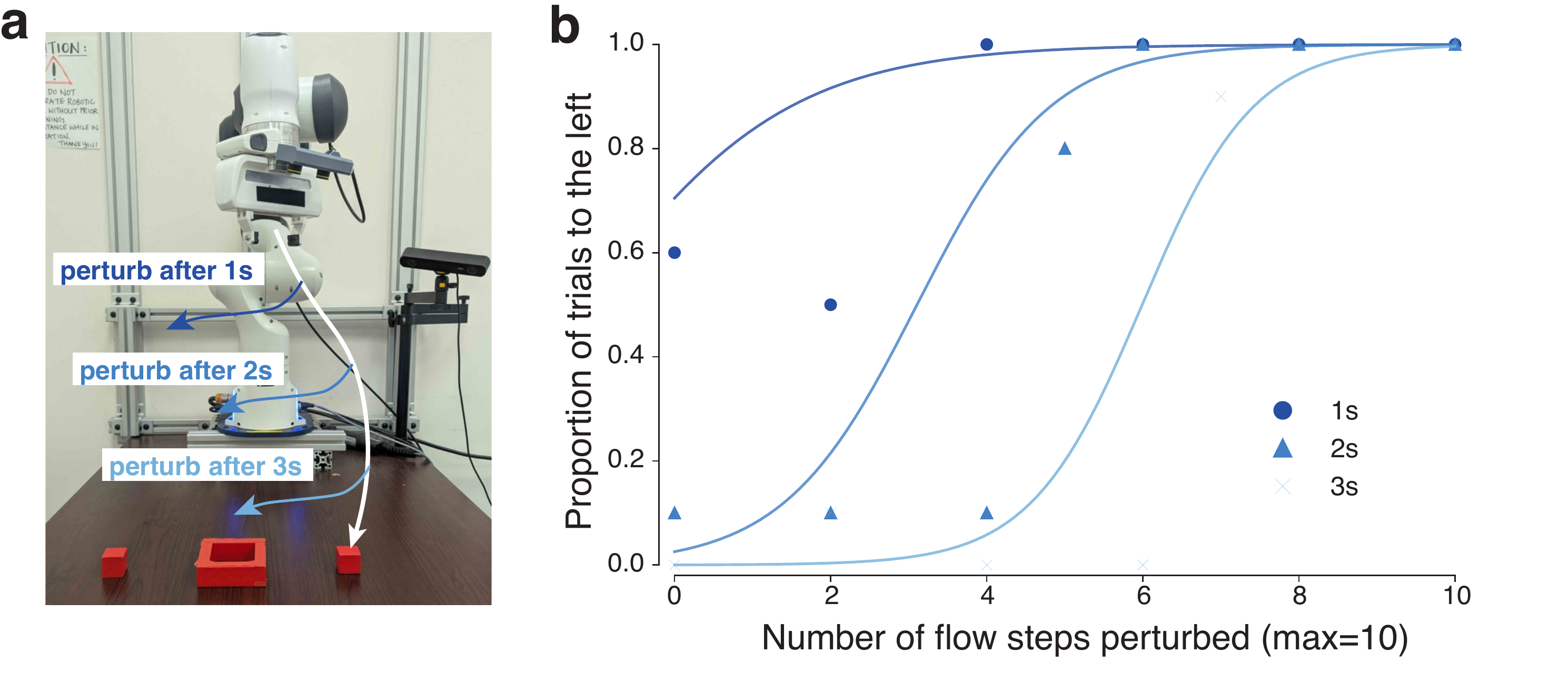}
    \caption{Perturbing the flow ODE integration.
    \textbf{a}, We perturb the flow process $1$~s, $2$~s, or $3$~s into the trajectory by modifying the flow.
    \textbf{b}, As more of the flow steps are perturbed, the policy more easily ``changes its mind'' to reach toward the left block.
    }
    \label{fig:fig6}
    \vspace{-\intextsep}
\end{wrapfigure}

%% file: app_teleop.tex
\section{Flow control and teleoperation}
\label{app:teleop}

\begin{wraptable}{r}{0.35\textwidth}
  \vspace{-\intextsep}
  \begin{tabular}{lcc}
    \toprule
    \textit{Time} [s] ($\downarrow$) & \textbf{Marker} & \textbf{Stack} \\
    \midrule
    Flow   & \textbf{13.30} & 35.21 \\
    Na\"ive  & 24.84 & 42.92 \\
    Expert & 15.39 & \textbf{30.97} \\
    \bottomrule
  \end{tabular}
  \caption{Trial time for flow control (all users), and for na\"ive and expert teleoperators (median split, $n{=}8$ each).}
  \label{tab:flow_vs_teleop}
  \vspace{-\intextsep}
\end{wraptable}

When we split operators by teleoperation skill (median split on teleoperation speed, $n{=}8$ each), flow control significantly sped up the na\"ive group (Wilcoxon signed-rank test, $p<0.05$ on Marker-in-Bowl, $p<0.05$ on Cup-Stacking) but not the expert group ($p=0.17$, $p=0.97$).
Flow control sped up the trial time for na\"ive operators from a mean of $24.8$~s (Marker-in-Bowl) and $42.9$~s (Cup-Stacking) with teleoperation to $12.4$~s and $36.6$~s with flow control ($50.3\%$ and $14.7\%$ faster), toward the expert-teleoperation level ($15.4$~s and $31.0$~s), while experts -- already fast -- were similar (Table~\ref{tab:flow_vs_teleop}, times not statistically significantly different).
The across-user variance in completion time consequently collapsed, by $32\times$ on Marker-in-Bowl and $2.8\times$ on Cup-Stacking.
Flow control completion times were nearly independent of an operator's teleoperation skill.
This can be intuitively understood as flow control ``leveling the playing field'': operators who gain the most are those least skilled at teleoperation.

%% file: app_implementation.tex
\section{Implementation details}
\label{app:impl}

This appendix (1) describes implementation details for how a keypress becomes the initial condition (IC) of the flow ODE, (2) characterizes the injected ICs and the user interventions, and (3) provides details of the fine-tuning experiments.

\textbf{Keyboard to joint command.}
Six keys commanded the $\pm x$, $\pm y$, and $\pm z$ directions of the end effector.
We polled these keys at $50$~Hz.
Several keys could be held at once.
Each held axis contributed approximately $11$~cm/s to the Cartesian velocity $\v_\user$, with zero angular velocity. %
We then used the regularized inverse kinematics (IK) of the DROID controller \cite{Khazatsky2024-bw} to convert $\v_\user$ into seven joint velocity commands, $\x_\user = \mathrm{IK}(\v_\user) \in \R^{7}$.
The IK used a regularization weight of $10^{-2}$ and a nullspace gain of $0.025$, with joint-limit handling.
The joint velocity commands are in the normalized command units of the DROID controller, in which $1$ corresponds to $0.2$~rad per control step. %

\textbf{Joint command to IC.}
The action expert of $\pi_{0.5}$-DROID integrates an IC $\x_0 \in \R^{16 \times 32}$.
This IC comprises $T=16$ chunk steps of $32$-dimensional action tokens.
We set the action horizon of $\pi_{0.5}$-DROID to $16$ (the public \texttt{pi05\_droid} configuration uses $15$).
Each action token holds $7$ joint velocities, the gripper position, and $24$ padding dimensions.
When a key was held at a replanning step, we sampled $\x_0 \sim \N(\0, \I)$ and then set
\begin{equation}
    \x_0[\,0{:}\tau,\; 0{:}7\,] \leftarrow g\, \x_\user,
    \qquad g = 15, \quad \tau = 8,
    \label{eq:impl-ic}
\end{equation}
which writes the same $7$-vector to each of the first $8$ chunk steps.
We did not clip these values.
The remaining gripper dimension, the padding dimensions, and the last $8$ chunk steps remained samples of $\N(0,1)$.
When no key was held, $\x_0$ was an unmodified sample of $\N(\0,\I)$, so the unsteered policy was autonomous $\pi_{0.5}$-DROID.

In this study, the $\textrm{normalize}(\cdot)$ step of Algorithm~\ref{alg:flowcontrol} is multiplication by the fixed gain $g$.
This was the only transformation we applied to $\x_\user$. %
The gain scales $\x_\user$ toward the unit per-coordinate scale of $\N(\0,\I)$.
We empirically found $g=15$ produced reliable steering.
We set $\tau=8$ based on the results of Section~\ref{subsec:steer} (Figure~\ref{fig:fig3}d).

\textbf{Integration and execution.}
We integrated the flow ODE with $K=10$ Euler steps.
We then kept the first $8$ dimensions of each action token, clipped to $[-1,1]$, and executed them at $15$~Hz.
The policy replanned after $8$ executed actions ($8/15 = 0.53$~s).
We aggregate constants for the flow control experiment in Table~\ref{tab:impl-constants}.

\begin{table}[h]
\centering
\small
\begin{tabular}{lc@{\hskip 2em}lc}
\toprule
Key polling rate & $50$~Hz & Control rate & $15$~Hz \\
Cartesian command per held axis & $\approx 11$~cm/s & Chunk length $T$ & $16$ \\
Gain $g$ & $15$ & Executed actions per chunk & $8$ \\
Steered chunk steps $\tau$ & $8$ & Replanning period & $0.53$~s \\
Steered dimensions per chunk step & $7$ (arm joints) & Euler steps $K$ of the flow ODE & $10$ \\
Overwritten IC coordinates & $56$ of $512$ & Clipping of executed actions & $[-1,1]$ \\
\bottomrule
\end{tabular}
\par\vspace{4pt}
\caption{Constants of the flow control implementation on $\pi_{0.5}$-DROID.}
\label{tab:impl-constants}
\end{table}

\textbf{Structural assumptions.}
As implemented, flow control assumes (i) a deterministic or low-noise action sampler whose IC lives in the action space (Appendix~\ref{app:samplers}); (ii) an IC prior of known scale; (iii) an action parameterization in which a low-dimensional command has a meaning, here joint velocities from IK; and (iv) action chunks that are replanned often enough for a command to take effect.
We tested flow control on a single VLA family ($\pi_{0.5}$ \cite{Intelligence2025-ht}), on a physical Franka Panda robot arm and in the LIBERO simulation \cite{Liu2023-libero} (Appendix~\ref{app:libero}).

\subsection{Statistics of the injected initial conditions}
\label{app:ic-stats}

The injected ICs $g\,\x_\user$ had a median magnitude of $0.46\sigma$ ($\sigma=1$ for $\N(0, 1)$), with $96\%$ within $2\sigma$.
Near kinematic singularities, the IK produced much larger joint velocity commands, with the 99th percentile reaching $6.2\sigma$.
To measure whether the generated actions stay close to the VLA's own actions, we performed additional experiments in Appendix~\ref{app:onpolicy}.

\subsection{Intervention statistics}
\label{app:interventions}

How often did users provide flow control inputs, and for how long?
To answer this question, we analyzed the flow control trials of the user study.

The user study comprised $314$ valid flow control trials.
Of these, $159$ were Marker-in-Bowl trials ($158$ successful) and $155$ were Cup-Stacking trials ($136$ successful).
We excluded the remaining $6$ of the $320$ scheduled trials, which were skipped or invalidated by technical failures (for example, a crash of the control script).
The success rates in Table~\ref{tab:auto_vs_flow} are pooled over these valid trials.
We computed the statistics in this appendix and in Appendices~\ref{app:ic-stats} and~\ref{app:onpolicy} from the trajectory logs of $290$ trials ($144$ Cup-Stacking and $146$ Marker-in-Bowl) from $15$ participants.

Keypresses were active on $17.9\%$ of time steps (range across the $15$ participants: $8.3$--$30.3\%$).
Users made $15.2$ keypress events per trial on average.
The median keypress hold duration was $0.13$~s (two control steps) and the mean duration was $0.37$~s.

Because a command reaches the VLA only at a replanning step, we also counted steered replans.
We found that $3{,}084$ of $17{,}475$ action chunks ($17.6\%$) were generated from a steered IC. %
At each of these replans, we computed the cosine similarity between the command $\x_\user$ and the joint-velocity dimensions of the first action of the chunk generated from that IC.
We used the action logged before clipping.
The median cosine similarity was $0.76$, and it was positive for $98.0\%$ of steered chunks.

\subsection{Fine-tuning details}
\label{app:finetune}

\textbf{Real robot.}
We fine-tuned $\pi_{0.5}$-DROID on $60$ successful flow control episodes of Cup-Stacking ($21{,}611$ time steps).
We recorded these episodes in a separate session after the user study, and kept only successful episodes.
A key was held on $14.3\%$ of these time steps.
Further, $363$ of the $2{,}724$ action chunks ($13.3\%$) were generated from a steered IC.
Training was standard behavior cloning on the executed actions.
We saved the keypresses in the raw trajectory logs, but they were not part of the training set.

We fine-tuned with low-rank adaptation (LoRA) \cite{Hu2022-lora}. %
We initialized from $\pi_{0.5}$-DROID, reused its normalization statistics, and did not use an exponential moving average of the weights. %
In both the fine-tuning episodes and the evaluation trials, we placed the three cups in one of five configurations: in a straight line, along one of two diagonals, or in one of two V shapes.
We did not control the distances between the cups, which varied from trial to trial.
We evaluated the fine-tuned policy, $\pi_{0.5}$-FC, on $25$ autonomous trials.
We evaluated $\pi_{0.5}$-DROID on $25$ autonomous Cup-Stacking trials ($30$ for Marker-in-Bowl).
The reported success rates are fractions of these trials.

\textbf{Simulation replication.}
To assess how performance compares when fine-tuning $\pi_{0.5}$ with flow control vs autonomous successes, we performed a fine-tuning experiment in LIBERO \cite{Liu2023-libero} with $\pi_{0.5}$-LIBERO on tasks 60 and 12 of LIBERO-90 (Appendix~\ref{app:libero}).
We chose these two tasks because they have moderate baseline performance with $\pi_{0.5}$-LIBERO, indicating the policy is competent enough that flow control steering could increase performance.
If the policy achieved zero performance, flow control would be unlikely to drive the policy toward actions that complete the task (since the policy rarely does), and if the policy achieved close to $100\%$ performance, there would be little headroom to measure performance improvement.
In this replication, we added a control dataset and held-out evaluation layouts.
We collected episodes from initial states $0$ to $34$ of each task, in up to two passes with different noise seeds, under two conditions.
The first condition was flow control driven by a scripted controller, which used privileged simulator state to select the key ($\tau=5$, injected magnitude $\sigma=1.5$).
The second condition was the unsteered policy.
For each condition, the first $30$ successful episodes per task formed a $60$-episode training set.

\begin{figure}[t!]
\centering
\includegraphics[width=0.7\linewidth]{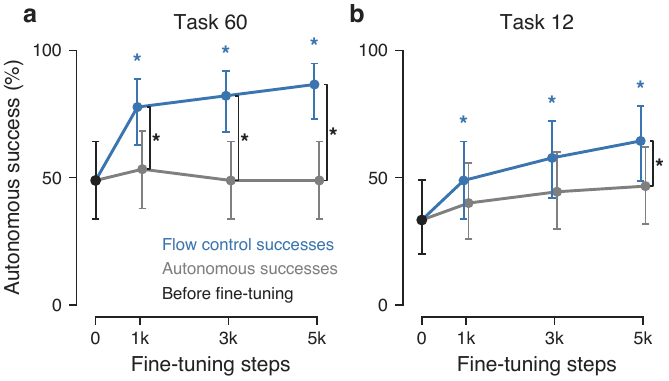}
\caption{Autonomous success of $\pi_{0.5}$-LIBERO on held-out initial states after fine-tuning on $60$ flow control successes (blue) or on $60$ of the policy's own successes (gray).
\textbf{a}, Task 60. \textbf{b}, Task 12.
The black point is the policy before fine-tuning.
Error bars are $95\%$ binomial confidence intervals ($45$ episodes per point).
A blue star marks a significant difference from the policy before fine-tuning.
A bracket with a star marks a significant difference between the two fine-tuned policies at that checkpoint ($p<0.05$, exact McNemar tests on paired episodes).
The final checkpoint was saved at step $4{,}999$.}
\label{fig:simft}
\end{figure}

We fine-tuned $\pi_{0.5}$-LIBERO on each set with LoRA \cite{Hu2022-lora} for $5{,}000$ steps.
We evaluated each checkpoint without steering on the held-out initial states $35$ to $49$ with three noise seeds, giving $45$ episodes per task (Figure~\ref{fig:simft}).
Every checkpoint was evaluated on the same initial states and noise seeds, so we compared conditions with exact McNemar tests on paired episodes.

Across both tasks, fine-tuning on flow control successes raised autonomous success from $41.1\%$ to $75.6\%$ at the final checkpoint ($p<0.01$).
In contrast, fine-tuning on autonomous successes reached $47.8\%$, which did not differ significantly from the policy before fine-tuning ($p=0.15$).
The difference between the two fine-tuned policies was significant at every checkpoint ($p < 0.01$).
On task 60, fine-tuning on flow control successes exceeded both the policy before fine-tuning and fine-tuning on autonomous successes at every checkpoint (Figure~\ref{fig:simft}a).
On task 12, it exceeded the policy before fine-tuning at every checkpoint, and exceeded fine-tuning on autonomous successes at the final checkpoint ($p< 0.01$; Figure~\ref{fig:simft}b).
In summary, fine-tuning on successes collected with flow control improved the autonomous policy, whereas fine-tuning on the policy's own successes did not significantly improve it.

%% file: app_mod_simulation.tex
\section{Simulation experiments in LIBERO}
\label{app:libero}

We report supporting experiments in the LIBERO simulation benchmark \citep{Liu2023-libero} with the public $\pi_{0.5}$-LIBERO checkpoint \citep{Intelligence2025-ht}.
In these experiments, we evaluate steered vs unsteered rollouts in LIBERO, quantify how ``on policy'' flow control actions are, and demonstrate that a VLM can steer the VLA.

\subsection{Setup and steering results}

\textbf{Setup.}
The $\pi_{0.5}$-LIBERO checkpoint generates action chunks of $10$ steps $\times$ $7$ dimensions, comprising $6$ Cartesian end effector deltas and the gripper.
It generates each action chunk from an initial condition (IC) of shape $10\times32$ in the normalized, zero-padded action space.
We replanned every $5$ steps at LIBERO's $20$~Hz control rate.
LIBERO-90 is held out from the checkpoint's fine-tuning data.
We used two LIBERO-90 tasks on which the unsteered policy has moderate success: task 60 (``pick up the black bowl on the left and put it in the tray'') and task 12 (``put the black bowl at the back on the plate'').
As previously described in Appendix~\ref{app:finetune}, we chose these tasks because $\pi_{0.5}$-LIBERO achieves moderate baseline performance.
If the policy achieved zero performance, flow control would be unlikely to drive the policy toward actions that complete the task (since the policy rarely does), and if the policy achieved close to $100\%$ performance, there would be little headroom to measure performance improvement.

\textbf{Scripted controller and injection.}
In these experiments, we replaced the user with a scripted controller.
At every replan, the scripted controller held the direction key for the dominant axis of the vector from the end effector to a target.
Before the grasp, the target was the task object.
After the grasp, the target was a point $12$~cm above the goal region.
When a key was held, we modified the first $\tau=5$ of the $10$ chunk steps of the IC.
We set the key's translation dimension to magnitude $\sigma=1.5$, with the sign given by the key's direction, and the other five translation and rotation dimensions to $0$.
We did not modify the gripper or padding dimensions.

\textbf{Results.}
We evaluated $50$ trials per task, using each task's $50$ fixed initial states.
Unsteered and flow control episodes shared the same initial state and noise stream.
Because outcomes are paired, we evaluated statistics using exact McNemar tests.
On task 60, flow control raised success from $27/50$ to $47/50$ ($p<0.01$).
On task 12, it raised success from $23/50$ to $33/50$ ($p<0.01$).

\subsection{Closeness to the VLA's own actions}
\label{app:onpolicy}
\label{app:baselines}

We quantify how ``on policy'' flow control actions are by calculating how close flow control action chunks are to the unsteered policy's own chunks.
We measured this by evaluating the BC (flow matching) loss of the frozen policy in LIBERO for flow control actions, as well as a nearest-neighbor distance.

\textbf{BC loss.}
Is the BC loss of flow control chunks similar to that of the policy's own chunks?
To answer this question, we scored chunks using the loss that the policy was trained on.
For a chunk $\a$ at observation $\c$, we computed the BC loss of the frozen $\pi_{0.5}$-LIBERO,
\begin{equation}
    \ell(\a \mid \c) = \frac{1}{J}\sum_{j=1}^{J} \frac{1}{D}\left\Vert \v^\theta_{t_j}\!\big((1-t_j)\,\eps_j + t_j\,\a \mid \c\big) - (\a - \eps_j) \right\Vert^2,
    \label{eq:bcloss}
\end{equation}
which is averaged over the $D=320$ entries of the padded chunk and over $J=32$ draws of $(t_j, \eps_j)$.
These draws were fixed and shared by all chunks.
We averaged over $D=320$ entries because this is the loss used during training in the \texttt{openpi} repository.
However, restricting to the $D=70$ entries ($10$ steps $\times$ $7$ dimensions) yielded the same conclusions.

On LIBERO-90 tasks 60 and 12, we scored $400$ chunks per task from each of four populations.
\emph{Unsteered} chunks are the chunks returned by the policy in unsteered rollouts.
\emph{Flow control} chunks are the chunks returned at the steered replans of flow control rollouts.
\emph{Action blending} adds the user command to the policy's action chunk instead of writing it into the IC.
In LIBERO, it added a fixed offset to the executed action on the axis of the scripted controller's key and clipped the result to $[-1,1]$.
Action blending chunks are the chunks executed in a separate set of action blending rollouts on the same tasks.
\emph{User command} chunks are the chunks that executing the command directly would send.

\begin{figure}[t!]
\centering
\includegraphics[width=0.8\linewidth]{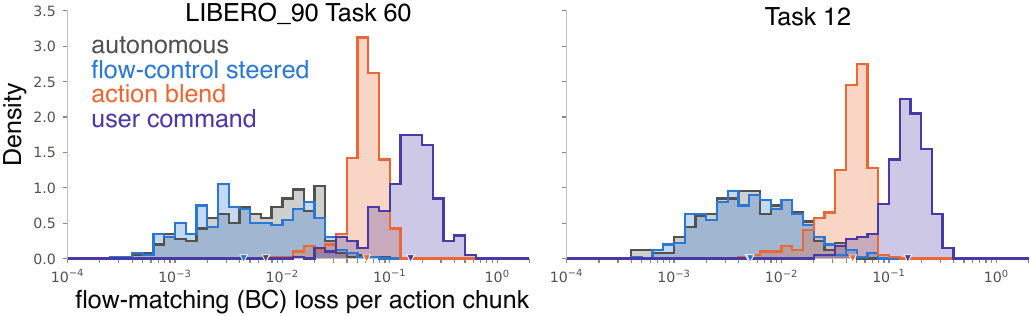}
\caption{Distribution of the BC loss of Equation~\eqref{eq:bcloss} under the frozen $\pi_{0.5}$-LIBERO on LIBERO-90 task 60 (left) and task 12 (right), for $400$ action chunks per population (logarithmic horizontal axis; triangles mark medians).
Gray: unsteered (``autonomous'') chunks; blue: flow control chunks; orange: chunks executed by action blending; purple: the user command executed directly.}
\label{fig:bcloss}
\end{figure}

\begin{table}[b!]
\centering
\small
\begin{tabular}{@{}lcccccc@{}}
\toprule
 & \multicolumn{3}{c}{Task 60} & \multicolumn{3}{c}{Task 12} \\
\cmidrule(lr){2-4}\cmidrule(l){5-7}
Population & Median & Ratio & In band & Median & Ratio & In band \\
\midrule
Unsteered & $0.0070$ & $1.00\times$ & $90\%$ & $0.0050$ & $1.00\times$ & $90\%$ \\
Flow control & $0.0044$ & $0.62\times$ & $88\%$ & $0.0051$ & $1.03\times$ & $91\%$ \\
Action blending & $0.0608$ & $8.68\times$ & $4\%$ & $0.0460$ & $9.21\times$ & $14\%$ \\
User command & $0.1562$ & $22.3\times$ & $2\%$ & $0.1502$ & $30.1\times$ & $0\%$ \\
\bottomrule
\end{tabular}
\par\vspace{4pt}
\caption{BC loss of the four chunk populations of Figure~\ref{fig:bcloss} ($n=400$ chunks per population per task, scored in one run per task).
Ratio: median relative to the unsteered median, computed from unrounded medians.
In band: fraction of chunks between the 5th and 95th percentiles of the unsteered losses.}
\label{tab:sim-bcloss}
\end{table}

We found that flow control chunks overlapped the unsteered distribution (Figure~\ref{fig:bcloss}, Table~\ref{tab:sim-bcloss}).
The median BC loss of flow control chunks was $0.62\times$ (task 60) and $1.03\times$ (task 12) that of unsteered chunks.
In contrast, it was $8.68\times$ and $9.21\times$ for action blending, and $22.3\times$ and $30.1\times$ for the user command.
Further, $88\%$ and $91\%$ of flow control chunks were between the 5th and 95th percentiles of the unsteered losses, compared to $4\%$ and $14\%$ for action blending and $2\%$ and $0\%$ for the user command.
On task 60, flow control chunks had lower loss than unsteered chunks (Wilcoxon rank sum test, $p<0.01$), and on task 12, flow control and unsteered chunks did not differ detectably ($p=0.69$).

\textbf{Nearest-neighbor (NN) distances.}
On the real robot, we asked how far flow control chunks are from the chunks that $\pi_{0.5}$-DROID generates on its own.
For a reference distribution of actions, we used the chunks of autonomous $\pi_{0.5}$-DROID rollouts on the same task.
We computed the Euclidean distance from each user-study chunk to its nearest neighbor in the reference.
As a baseline, we used the user-study chunks generated with no key held, which are the VLA's own output.
The median NN distance of flow control chunks was $1.36\times$ (Cup-Stacking) and $1.08\times$ (Marker-in-Bowl) that of this baseline.
With the teleoperation demonstrations that the same participants recorded for the same task as the reference, these ratios were $1.44\times$ and $1.05\times$.
On LIBERO-90 tasks 60 and 12, with a random half of the unsteered chunks as the reference, the median NN distance of flow control chunks was $1.07\times$ (task 60) and $1.01\times$ (task 12) that of the held-out unsteered chunks.

\textbf{Why not steer with language?}
We can also steer a frozen VLA through language corrections.
Its actions are ``on policy'' by construction, because the IC is unmodified and only the prompt changes.
Why not use them?
We found that the language corrections we tested did not steer the VLA.
On the real robot, we replayed the $363$ logged intervention states of the $60$ flow control episodes used for fine-tuning (Section~\ref{subsec:finetune}) through $\pi_{0.5}$-DROID offline.
We compared the base prompt with three corrections.
These were the task instruction followed by ``move to the $\langle$direction$\rangle$'', the task instruction followed by ``no, to the $\langle$direction$\rangle$'', and a correction-only prompt.
The median projection of the first $8$ actions of the returned chunk onto the unit command direction was between $-0.07$ and $-0.01$ action units per step for these four prompts, compared to $+0.33$ for flow control replayed at the same states.
This replay measures instantaneous compliance, not closed-loop task success.
In LIBERO, we appended ``, the one on the left'' (task 60) or ``, the one at the back'' (task 12) to the instruction and ran the policy in closed loop.
It succeeded on $28/50$ and $24/50$ episodes, which did not differ significantly from the unsteered policy in the same runs ($27/50$ and $21/50$; $p=1.00$ and $p=0.38$).
In contrast, flow control reached $49/50$ and $33/50$ in these runs.
We did not test other prompts beyond these.
This is consistent with \citet{Chen2026-steerable}, who argue that a VLA cannot follow motion-level language commands such as ``move left'' unless it is trained on such commands.

\subsection{Flow control steering with vision-language models}
\label{app:vlm}

Flow control only requires a direction, so the command does not have to come from a human user.
Can a VLM provide the steering command for flow control?
To answer this question, we replaced the scripted controller of Appendix~\ref{app:libero} with a controller built from VLMs.

\textbf{Setup and VLM controller.}
We used $\pi_{0.5}$-LIBERO with the flow control interface of Appendix~\ref{app:libero}.
The controller used two open-weight models, which we served locally and did not fine-tune.
We used Qwen3.6-35B-A3B for reasoning and Molmo2-ER for pointing. %
The controller computed its target in two stages.
In the first stage, the reasoning model received the task instruction, the end effector position, the gripper state, up to three previous images paired with their corresponding commands, and the current $256 \times 256$ front camera image.
It answered with the object the gripper must act on next, the object it must be carried to, and one of four phases (reach, grasp, transport, place).
In the second stage, we asked the pointing model to ``Point to the $\langle$object$\rangle$.''
We lifted the returned pixel to a 3D point with the depth image.
Before a grasp, the target was this point plus a $4$~cm hover offset.
After a grasp, the target was the destination plus $12$~cm.

At each replan, the command was the unit direction $\hat{\mathbf{d}}$ from the end effector to the target.
We tested two ways of writing this direction into the IC.
The first, \textit{continuous}, writes $\sigma \hat{\mathbf{d}}$ on the three translation dimensions.
The second, \textit{six-key}, first snaps $\hat{\mathbf{d}}$ to its dominant axis, and therefore uses exactly the six directions of the keyboard.
We queried the VLM every $5$ action chunks, or when the grasp state changed.
Like the scripted controller, the VLM controller was given the simulator's grasp state of the target object, so that both knew when the object was held.

\begin{table}[b!]
  \centering
  \small
  \begin{tabular}{llcc}
    \toprule
    \textbf{Command source} & \textbf{Direction} & \textbf{Task 60} & \textbf{Task 12} \\
    \midrule
    None (unsteered) & -- & $28/50$ & $22/50$ \\
    Scripted controller & six-key & $47/50$ & $32/50$ \\
    VLM & continuous & $46/50$ & $34/50$ \\
    VLM & six-key & $47/50$ & $30/50$ \\
    \bottomrule
  \end{tabular}
  \par\vspace{4pt}
  \caption{Task success of $\pi_{0.5}$-LIBERO steered with flow control by different command sources ($50$ paired episodes per cell, all cells run together).
  Task 60 is ``pick up the black bowl on the left and put it in the tray'' and task 12 is ``put the black bowl at the back on the plate.''}
  \label{tab:vlm}
\end{table}

\textbf{Results.}
We evaluated the performance of the VLM on $50$ initial states and noise seeds, using the exact two-sided McNemar test on paired episodes (Table~\ref{tab:vlm}).
On task 60, the VLM reached $46/50$ with continuous directions and $47/50$ with six-key directions.
These did not differ from the scripted controller ($47/50$, $p = 1.0$ for both), and exceeded the unsteered policy ($28/50$, $p<0.05$ for both controllers).
On task 12, the VLM reached $34/50$ and $30/50$.
These also did not differ from the scripted controller ($32/50$, $p = 0.5$ for both), and exceeded the unsteered policy ($22/50$, $p<0.05$ for both controllers).
The VLM therefore steered $\pi_{0.5}$-LIBERO as well as the scripted controller on both tasks.

\subsection{Comparison with Flow Reversal Steering}
\label{app:frs}

Flow Reversal Steering (FRS) \cite{Tang2026-frs} is concurrent work that also steers flow matching policies through their noise.
FRS builds a reference action chunk from the command.
It then integrates the flow ODE in reverse to map the reference to noise, and integrates forward from that noise.
In contrast, flow control writes the command into the IC and integrates forward once, with the policy's unmodified sampler.

\textbf{FRS depends on the integration step size.}
When the reverse and forward passes use the same observation and an exact integrator, the forward pass inverts the reverse pass and returns the reference chunk unchanged.
The action that FRS executes therefore depends on the step size of the integrator.
With $N$ reverse and $N$ forward Euler steps, a smaller step size (a larger $N$) brings the executed chunk closer to the reference chunk, which is the user command played directly.
As we showed in Appendix~\ref{app:onpolicy}, the user command can often be OOD with respect to the VLA's actions. %

\textbf{Implementation.}
We evaluated FRS with our own implementation based on the description in \citet{Tang2026-frs}.
In our implementation, the reference chunk moved in a straight line along the commanded direction on all $10$ time steps, with no rotation and the previous gripper command.
The reference had a fixed per-step magnitude of $0.55$ action units.
This is close to the typical per-step magnitudes of the unsteered policy on tasks 60 and 12 ($0.59$ and $0.53$, respectively).
We inverted the full chunk with $N$ reverse Euler steps and then sampled with $N$ forward Euler steps.
Here, $N = 10$ corresponds to the step size used by \citet{Tang2026-frs}.
We used $\pi_{0.5}$-LIBERO with the previously described scripted controller on tasks 60 and 12.

\begin{figure}[t!]
\centering
\includegraphics[width=0.9\linewidth]{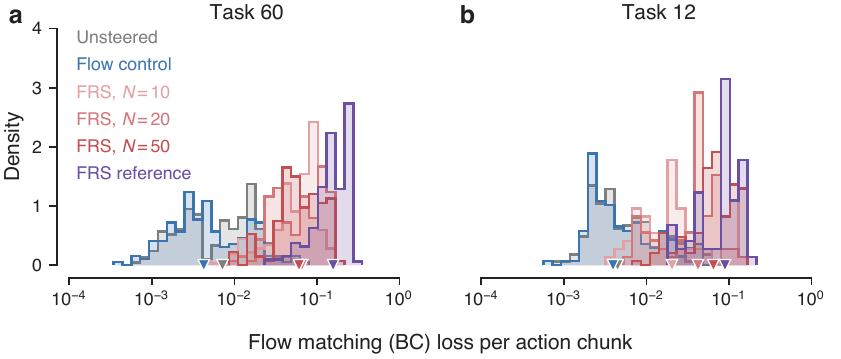}
\caption{BC loss of the frozen $\pi_{0.5}$-LIBERO on executed action chunks, on LIBERO-90 task 60 (\textbf{a}) and task 12 (\textbf{b}), for the replans at which the scripted controller issued a command ($20$ episodes per method and task; logarithmic horizontal axis; triangles mark medians).
Gray: unsteered chunks. Blue: flow control. Red, from light to dark: FRS (our implementation) with $N = 10$, $20$, and $50$. Purple: the reference chunk that FRS inverts (from the $N = 10$ runs).}
\label{fig:frs-bcloss}
\end{figure}

\textbf{Results.}
We first measured how close the executed chunk was to the reference chunk.
In closed loop, the relative distance between them was $0.50$, $0.32$, and $0.13$ at $N = 10$, $20$, and $50$, where $0$ means that the reference is executed unchanged.
As the step size decreased, FRS approached executing the reference.

We then evaluated the BC loss (Equation~\eqref{eq:bcloss}) of the frozen VLA on the executed chunks, as in Appendix~\ref{app:onpolicy}.
We used the replans at which the scripted controller issued a command, in $20$ episodes per method and task (Figure~\ref{fig:frs-bcloss}, Table~\ref{tab:frs-bcloss}).
The median BC loss of flow control chunks was $0.59\times$ (task 60) and $0.88\times$ (task 12) that of unsteered chunks.
These ratios differ from those of Appendix~\ref{app:onpolicy} because they come from different rollouts and fewer chunks ($20$ episodes per method and task); the differences are within sampling variability.
At all step sizes, we found that FRS chunks had significantly larger BC loss than unsteered chunks ($p<0.01$, Wilcoxon rank sum test).
For FRS chunks with $N = 10$, the median BC loss was $9.1\times$ (task 60) and $4.5\times$ (task 12) that of unsteered chunks, and $21.9\times$ (task 60) and $19.9\times$ (task 12) for reference chunks.
Further, $88\%$ of flow control chunks on both tasks were between the 5th and 95th percentiles of the unsteered losses.
In contrast, $15\%$ (task 60) and $66\%$ (task 12) of FRS chunks with $N = 10$ were in this range, compared to $17\%$ and $27\%$ with $N = 20$, and $10\%$ and $14\%$ with $N = 50$.
For reference chunks, $0\%$ and $7\%$ were in this range.
FRS chunks were therefore further from the VLA's actions than flow control chunks at every step size we tested (Figure~\ref{fig:frs-bcloss}).
We therefore found that in these two simulated tasks, FRS generated actions closer to the commanded input than to the VLA's own actions.
We note that this does not necessarily correlate to success on the task; especially when the policy is not excellent, FRS actions that more closely match the user actions may increase task performance.
In contrast, flow control aims to generate an action close to the VLA's actions from a crude user action.

\begin{table}[h]
\centering
\small
\begin{tabular}{@{}lcccccccc@{}}
\toprule
 & \multicolumn{4}{c}{Task 60} & \multicolumn{4}{c}{Task 12} \\
\cmidrule(lr){2-5}\cmidrule(l){6-9}
Population & $n$ & Median & Ratio & In band & $n$ & Median & Ratio & In band \\
\midrule
Unsteered & $196$ & $0.0072$ & $1.00\times$ & $90\%$ & $152$ & $0.0045$ & $1.00\times$ & $89\%$ \\
Flow control & $180$ & $0.0043$ & $0.59\times$ & $88\%$ & $128$ & $0.0039$ & $0.88\times$ & $88\%$ \\
FRS, $N = 10$ & $296$ & $0.0656$ & $9.13\times$ & $15\%$ & $68$ & $0.0204$ & $4.55\times$ & $66\%$ \\
FRS, $N = 20$ & $128$ & $0.0658$ & $9.16\times$ & $17\%$ & $67$ & $0.0421$ & $9.38\times$ & $27\%$ \\
FRS, $N = 50$ & $124$ & $0.0607$ & $8.45\times$ & $10\%$ & $136$ & $0.0647$ & $14.4\times$ & $14\%$ \\
FRS reference & $296$ & $0.1572$ & $21.9\times$ & $0\%$ & $68$ & $0.0896$ & $19.9\times$ & $7\%$ \\
\bottomrule
\end{tabular}
\par\vspace{4pt}
\caption{BC loss of the chunk populations of Figure~\ref{fig:frs-bcloss} ($20$ episodes per method and task).
$n$: number of chunks.
Ratio: median relative to the unsteered median, computed from unrounded medians.
In band: fraction of chunks between the 5th and 95th percentiles of the unsteered losses.
FRS reference: the reference chunk that FRS inverts (from the $N = 10$ runs).}
\label{tab:frs-bcloss}
\end{table}